# Machine Learning Based Multimodal Neuroimaging Genomics Dementia Score for Predicting Future Conversion to Alzheimer's Disease

Ghazal Mirabnahrazam [a,1], Da Ma [b,a], Sieun Lee [a,g], Karteek Popuri [a], Hyunwoo Lee [c], Jiguo Cao [d], Lei Wang [e], James E Galvin f, Mirza Faisal Beg [a,*], and the Alzheimer's Disease Neuroimaging Initiative [2]

[a] *School of Engineering, Simon Fraser University, Burnaby, BC, Canada*
[b] *School of Medicine, Wake Forest University, Winston-Salem, NC, USA*
[c] *Division of Neurology, Department of Medicine, University of British Columbia, Vancouver, BC, Canada*
[d] *Department of Statistics and Actuarial Science, Simon Fraser University, Burnaby, BC, Canada*
[e] *Psychiatry and Behavioral Health, Ohio State University Wexner Medical Center, Columbus, OH, USA*
[f] *Comprehensive Center for Brain Health, Department of Neurology, University of Miami Miller School of Medicine, Miami, FL, USA*
[g] *Mental Health & Clinical Neurosciences, School of Medicine, University of Nottingham, Nottingham, United Kingdom*

---

## Abstract

**Background:** The increasing availability of databases containing both magnetic resonance imaging (MRI) and genetic data allows researchers to utilize multimodal data to better understand the characteristics of dementia of Alzheimer's type (DAT).

**Objective:** The goal of this study was to develop and analyze novel biomarkers that can help predict the development and progression of DAT.

**Methods:** We used feature selection and ensemble learning classifier to develop an image/genotype- based DAT score that represents a subject's likelihood of developing DAT in the future. Three feature types were used: MRI only, genetic only, and combined multimodal data. We used a novel data stratification method to better represent different stages of DAT. Using a pre-defined 0.5 threshold on DAT scores, we predicted whether or not a subject would develop DAT in the future.

**Results:** Our results on Alzheimer's Disease Neuroimaging Initiative (ADNI) database showed that dementia scores using genetic data could better predict future DAT progression for currently normal control subjects (Accuracy=0.857) compared to MRI (Accuracy=0.143), while MRI can better characterize subjects with stable mild cognitive impairment (Accuracy=0.614) compared to genetics (Accuracy=0.356). Combining MRI and genetic data showed improved classification performance in the remaining stratified groups.

**Conclusion:** MRI and genetic data can contribute to DAT prediction in different ways. MRI data reflects anatomical changes in the brain, while genetic data can detect the risk of DAT progression prior to the symptomatic onset. Combining information from multimodal data in the right way can improve prediction performance.

---

*Corresponding author: Mirza Faisal Beg, PhD, P.Eng., Michael Smith Foundation for Health Research Scholar, School of Engineering Science, Simon Fraser University, ASB 8857, 8888 University Drive, Phone: (778) 782-5696, Website: http://www2.ensc.sfu.ca/~mfbeg/

Email addresses: ghazal_mirabnahrazam@sfu.ca (Ghazal Mirabnahrazam), da_ma@sfu.ca (Da Ma), sieun.lee@nottingham.ac.uk (Sieun Lee), kpopuri@sfu.ca (Karteek Popuri), hyunwoo.lee@ubc.ca (Hyunwoo Lee), jiguo_cao@sfu.ca (Jiguo Cao), Lei.Wang@osumc.edu (Lei Wang), jeg200@miami.edu (James E Galvin), faisal-lab@sfu.ca (Mirza Faisal Beg)

1 First Author

2 Data used in preparation of this article were obtained from the Alzheimer's Disease Neuroimaging Initiative (ADNI) database (http://adni.loni.usc.edu). As such, the investigators within the ADNI contributed to the design and implementation of ADNI and/or provided data but did not participate in analysis or writing of this report. A complete listing of ADNI investigators can be found at http://adni.loni.usc.edu/wp-content/uploads/how_to_apply/ADNI_ Acknowledgement_List.pdf

---

## INTRODUCTION

Alzheimer's Disease (AD), or Dementia of Alzheimer's type (DAT), is a progressive neurodegenerative condition characterized by psychiatric, cognitive and structural deteriorations, accounting for 60% to 80% of all dementia cases [1]. As there is no currently available cure, there is a substantial interest in finding biomarkers that can detect those at risk at early stage of the disease before the symptomatic onset. Data from various modalities have been obtained and analyzed in search of biomarkers that can reliably diagnose DAT at its early stages. For example, magnetic resonance imaging (MRI) is the most widely used data modality for identifying characteristic structural changes in the brain associated with DAT progression [2–6]. Genetic information is another modality that has been shown to be effective in predicting the likelihood of developing DAT even before pathological changes begin. A number of genetic risk factors have been found to be associated with DAT, among which the APOE-ε4 allele accounts for 20-25 percent of the cases [7]. Multiple genome-wide association studies (GWAS) have also demonstrated potential associations between single nucleotide polymorphisms (SNPs) and DAT [8–13]. At the time of writing this manuscript, 20 genes has been reported to be associated with AD, identified through GWAS, most of which are associated with moderate to small effect sizes [7].

MRI and genetic data have distinct properties that can contribute to the prediction of DAT progression. MRI data provides tissue level information and may reflect phenotype information about anatomical changes in the brain since the early stages of DAT, and genetic data provides molecular level information and may encode genotype information of probable DAT progression even in the absence of detectable brain changes. Combining information from genotype and phenotype data may reveal patterns that are not visible when working with individual modalities separately, allowing for more robust predictions. With the increasing availability of databases that contain both MRI and genetic data, such as the Alzheimer's Disease Neuroimaging Initiative (ADNI), multiple studies have explored the effects of integrating both modalities in DAT risk prediction, suggesting that combining the complementary information from both modalities can enhance the diagnosis performance [14–22]. However, existing image/genotype studies of DAT mainly focused on SNPs from previously known DAT-related genes [14–22]. Such approaches rely on existing knowledge and may reduce the chances of discovering novel genetic risk factors.

In this study, we address the above-mentioned potential limitations in the current research of genetic study for AD by using all available SNPs in the ADNI database to uncover potentially new genetic risk factors of DAT. We have designed a robust feature selection technique to address the high dimensionality of genetic data. We proposed an automated framework to achieve the prognosis of DAT by extracting and fusing information from both brain MRI and genetic data collected from subjects at various stages of the disease, and developed a novel image/genotype-based dementia score indicating the probability of a subject developing DAT in the future. We have investigated the effects of using data from MRI, genetic, and combined modalities on DAT prediction separately, and provided a detailed report on how each modality contributes to DAT diagnosis.

## METHODS

There are three main steps in the proposed framework: (1) data processing: a) brain MRI: segmenting brain tissue, parcellating brain structural regions and extracting volumetric features, b) SNP: quality control to remove subjects and SNPs with low quality data, (2) feature selection: performing association tests on each modality; and, (3) disease stage classification: training a machine learning network with the most discriminative features selected above, and then using for classification.

*Experimental data*

Brain MRI and genetic data used in preparation of this article were obtained from the publicly available Alzheimer's Disease Neuroimaging Initiative (ADNI) database (http://adni.loni.usc.edu). The ADNI was launched in 2003 as a public-private partnership, led by Principal Investigator Michael W. Weiner, MD. The primary goal of ADNI has been to test whether serial MRI, PET, other biological markers, and clinical and neuropsychological assessment can be combined to measure the progression of mild cognitive impairment (MCI) and early Alzheimer's disease (AD). In addition, ADNI aims to provide researchers with the opportunity to combine genetics with imaging and clinical data to help investigate mechanisms of the disease.

*Group stratification*

A total of 543 subjects from the first phase of ADNI (ADNI1) [23] who had both MRI and genetic data available were included in the study. We utilized a database stratification method focusing on the past, current and future clinical diagnosis of the subjects in the study [24]. This method divides the subjects into seven subgroups based on their screening and follow-up clinical diagnosis, in addition to their clinical diagnosis at the time of the MRI imaging visit. Each MRI image corresponds to a clinical diagnosis, and participants may receive multiple diagnoses based on their MRI images acquired during the study period. Participants' genetic information, on the other hand, remains constant over time. As a result, in this study, we only used each participant's baseline MRI data, as well as genetic information.

Based on the information available during the ADNI study period, each participant was assigned to one of the seven subgroups described below:

- **sNC** (stable NC): Subjects with a normal control (NC) diagnosis at baseline imaging visit whose diagnosis remained unchanged throughout the study window;
- **uNC** (unstable NC): Subjects with NC diagnosis at baseline imaging visit who progressed to MCI at a future timepoint in the study window;
- **pNC** (progressive NC): Subjects with NC diagnosis at baseline imaging visit who progressed to DAT at a future timepoint in the study window;
- **sMCI** (stable MCI): Subjects with MCI diagnosis at baseline imaging visit whose diagnosis remained unchanged throughout the study window;
- **pMCI** (progressive MCI): Subjects with MCI diagnosis at baseline imaging visit who progressed to DAT at a future timepoint in the study window;
- **eDAT** (early DAT): Subjects with DAT diagnosis at baseline imaging visit who received NC or MCI status at

an earlier screening (non-imaging) visit in the study window;

- **sDAT** (stable DAT): Subjects with DAT diagnosis at baseline imaging visit and earlier visits throughout the study window.

Subjects in the pNC, pMCI, eDAT, and sDAT subgroups are labelled DAT+, indicating that they follow a DAT trajectory and developed DAT during the study window. The sNC, uNC and sMCI subjects do not progress to DAT during the study period, hence are denoted as DAT-. The eDAT group has a small sample size (4 subjects), but it has been included in the study for the sake of completeness. The aim of this study was to predict a subject's future conversion to DAT based on its baseline MRI image and genetic data. Table 1 shows the demographic information for the ADNI subjects used in our experiments, as well as their disease progression subgroup stratification.

*Data processing*

*Genetic data processing*

Genotyping information of 757 ADNI1 subjects was downloaded in PLINK [25] format from the LONI Image Data Archive (https://ida.loni.usc.edu/). During the genotyping phase, 620,901 SNPs were obtained on the Illumina Human610-Quad BeadChip platform. APOE was genotyped separately during the study's screening phase [26]. Genomic quality control was conducted using the PLINK software and included the following steps:

- SNP-specific and subject-specific missingness rate check;
- Minor Allele Frequency (MAF) check;
- Hardy-Weinberg equilibrium (HWE) test;
- Gender check;
- Sibling pair identification and removal;
- Heterozygosity rate check;
- Population Stratification

The above procedure yielded 521,014 SNPs for 570 subjects. Our SNP data was then recoded to reflect the number of minor (second most common) alleles per person for each SNP. The categorical SNP features can obtain one of the following possible values: −1 for missing information; 0 for homozygous major alleles (2 major alleles); 1 for heterozygous alleles (1 minor and 1 major alleles); and 2 for homozygous minor alleles (2 minor alleles).

APOE-ε2, APOE-ε3, and APOE-ε4 were then added to SNPs. These three APOE alleles are also categorical, and each of them can obtain one of the following three values: −1 for missing information; 0 if the allele does not exist; and 1 if it does. In the remaining text of the manuscript, we refer to the combination of SNP and APOE features as the "genetic features" (521, 014 + 3 = 521, 017 features). Finally, we excluded subjects that had no diagnosis label available, leaving 543 subjects for our analysis.

*Brain MRI processing*

We used the FreeSurfer software (version 5.3) (http://surfer.nmr.mgh.harvard.edu/) to segment the T1-weighted baseline MRI images into the gray matter (GM), white matter (WM), and cerebrospinal fluid (CSF) [27] regions. Extensive manual quality control was then employed to correct the automated tissue segmentations according to the FreeSurfer guidelines. Following the QC step, we used Freesurfer's cortical [28] and subcortical [29] labeling pipelines and divided the GM and CSF tissue regions into 91 distinct regions.

We used a generalized linear model (GLM) framework introduced in our previous publication [30] to remove the individual heterogeneity due to sex, scanner field strength, scanner type, and total intracranial vault (TIV) to only retain differences due to AD-induced volume change. Following the data harmonization step, for each baseline image, we calculated the standardized residual value (w-score) from the measured AD-related volumes to be used as the features alongside the genetic features to train the machine-learning classifier for computing the DAT score. Details regarding calculating the w-score can be found in our previous publications [4, 30].

*Feature selection and DAT score computation via supervised ensemble learning*

To compute the proposed DAT score from the w-score brain volume features (MRI-based score), genetic features (genetic-based score), and the combination (MRI+genetic-based score), we used a two- step supervised classification model that combines multiple distinctly trained classifiers into a single, more robust classification model using the ensemble learning technique [31]. We have previously used this technique to develop a) fluorodeoxyglucose positron emission tomography (FDG-PET) imaging- based score [24] and b) MRI-based score [4] for early DAT detection and achieved the state-of-the-art performance.

In the first step, the model was used to select the most discriminative features from MRI and genetic data, and in the second step, the DAT score was computed using the fixed set of features selected before. In both steps, only subjects from the sNC (N=109) and sDAT (N=138) classes (247 subjects), the groups with the highest clinical diagnosis certainty, were used to train the model. The sDAT class represents the DAT+ group and the sNC class represents the DAT- group.

To avoid overfitting on the training data, the sub-bagging approach [32] was employed to randomly generate F=10 subsets of the training data with a sampling ratio of 0.8. To avoid class imbalance, we used stratified sampling to select the same number of subjects from each class (based on the class with the smaller sample size; here sNC), i.e., $N_{train} = 2 \times [0.8 \times 109] \approx 174$ subjects randomly selected from a total of 247 sNC and sDAT subjects in each of the F training subsets.

<u>*Step 1: Feature selection*</u>

There is an experimental relationship between the size of training data and the maximum number of features ($K_{max}$) that can be used to train a classifier in order to avoid the "curse of dimensionality" and minimize the risk of overfitting, which is that for each F subset of training data with $N_{train}$ number of samples, a maximum number of $K_{max} = N_{train} \times 2p(e)$ features is required to train the classifier, where p(e) is the probability of error [33, 34]. Our goal was to keep $p(e)$ as low as possible for all of our experiments while still having enough features to train the classifiers. To keep $p(e)$ below 5% when using either MRI or genetic data, $K_{max} = N_{train} \times 2 \times 0.05 = N_{train}/10$ [35]. Therefore, k = 174/10 ≈ 17 features were selected each time based on their effect size on the outcome.

To identify the most discriminative features, we performed statistic-based feature ranking and selection on MRI and genetic features separately, determining if each features has a significant relationship with the outcome (here, being on the DAT+ trajectory). Statistical-based feature selection methods are fast, however, in order to select the right algorithm, it is important to pay attention to the data type of both input and output variables [36]. Different statistical tests were selected for MRI and genetic data after extensive examination and with careful attention to their data type (i.e. continuous vs. categorical). Specifically, we applied Fisher's Exact test [37], a statistical significance test designed for the categorical data type that examines each feature individually and assigns an exact significance value to each feature, on 521,017 categorical genetic features and Welch's t-test [38] on 91 continuous w-score volume features, and for each feature type, we obtained F = 10 independent sets of k = 17 features with the largest effect sizes on the outcome. We used effect size rather than p-value to rank the significance of the features because p-values are affected by sample size and a statistically significant p-value may indicate that a large sample size was used rather than demonstrating an actual significant difference.

The features were then ranked based on their frequency of selection in the F subsets, and the first k = 17 most frequently selected features were chosen for the next step to ensure a strong association between the selected features and the disease pattern. In cases where features had similar selection frequency, they were deemed to be of equal importance. When necessary, a final set of 17 features was formed by random selection from equally important features. Finally, to investigate the combined effect of MRI and genetic data on the DAT score computation, we combined the selected unique features from both feature types (17 + 17 = 34 MRI+genetic features). Figure 1 illustrates the feature selection process for MRI and genetic data.

To evaluate the efficacy of our data-type-specific feature selection procedure, we compared our method with the Least Absolute Shrinkage and Selection Operator (LASSO) [40], which is a regression based feature selection algorithm that can be used on both categorical and continuous variables to select the most discriminative features. We replaced LASSO with Fisher's exact test for genetic data and Welch's t-test for MRI data to pick the most discriminative features in the same settings as before.

To support the value of the SNPs selected using our feature selection method, we trained our model with two sets of known AD-related SNPs as features. The first set includes 17 AD-related SNPs reported by Giri et al. [54], and the second set includes 17 SNPs from the top 10 AD-related genes reported in the Alzgene database (http://www.alzgene.org/). Table 2 contains information about the SNPs identified in the aforementioned studies. Not all SNPs listed in these two studies were available in the ADNI dataset.

*Step 2: DAT score computation*

We trained a probabilistic multi-kernel classifier, Variational Bayes Probabilistic Multi-Kernel Learning (VBpMKL) [41] on F training subsets containing 174 randomly selected sNC and sDAT subjects. The VBpMKL classifier performs hyperparameter tuning by applying different kernels [42] (e.g., Gaussian, second-order polynomial) to each feature space and learning the weight of each kernel for different features using the variational Bayesian approximation, and then outputs a probabilistic estimation to each class for each data. The kernels applied to each feature space in this study were linear, first-order polynomial, second-order polynomial, and third-order polynomial.

The above procedure was performed separately using the MRI, genetic or MRI+genetic features selected in Step 1. After training, each probabilistic kernel classifier outputted the probability pi ∈ [0 1], i = {1,...,F} that the input data belonged to the DAT+ class (1 − pi denotes the probability of DAT- membership). The final image/genotype-based DAT score (MRI DAT score: MRDATS, genetic DAT score: GENDATS, and MRI+genetic DAT score: MRGENDATS) was then defined as the average of all probabilistic predictions over F classifiers. The DAT score is scalar and can be viewed as a measure of similarity to the DAT-/DAT+ classes, i.e., a score close to 1 indicates similarity with the DAT+ and a score close to 0 reveals similarity with the DAT- class.

To avoid biased estimates, the DAT score values were calculated using the out-of-bag estimation method (using only the remaining 20% of the subjects in each of the F training subsets) [24]. The DAT score for a subject in sNC and sDAT groups (training groups) was calculated using only predictions from ensemble classifiers that did not have that subject in their training subset. We further evaluated the performance of the trained ensemble model on the remaining stratified subgroups (uNC, pNC, sMCI, pMCI, and eDAT; testing groups). The subjects belonging to these groups were unseen by the classifiers since they have not been included in the training process.

A threshold of 0.5 was used to create a diagnostic label of DAT- or DAT+ from the DAT score. Sensitivity, specificity, accuracy, and balanced accuracy were then obtained by comparing the label to the actual clinical diagnosis. The area under the curve was also calculated by scanning the threshold from 0 to 1 which is an indication of the separation of the class (DAT-/DAT+) histograms. To compare the group-wise differences, DAT score distribution and prediction accuracy were also obtained for each stratified group.

## RESULTS

*Salient feature selection for DAT score computation*

The final set of 17 features for each feature type was obtained by choosing the most frequently selected features in the F=10 classifiers. Figure 2 shows the frequency of selection for the entire set of features chosen at least once by the classifier ensemble. Features with a similar selection frequency were deemed equally important. For example, Figure 2, bottom row, shows that 12 genetic feature were chosen twice (20%) and thus were of equal importance. To create the final set of 17 features for genetic data, we randomly selected 7 of the 12 features chosen twice (20%) by the classifier ensemble, as well as the first 10 features chosen more than twice. The final set of features for each feature type is highlighted in Figure 2.

14 of the top 17 MRI features were chosen all the time (100%) by the classifiers in the ensemble. Table 3 includes information about the most discriminative MRI features determined by our feature selection method. These features include the volumetric measures of brain regions such as amygdala, hippocampus, entorhinal cortex, parahippocampal cortex and fusiform gyrus, which are all well-known biomarkers of DAT. These regions are consistent with many previous studies, which shows the effectiveness of our proposed method [4, 24]. The top selected genetic features are more scattered and have a smaller selection frequency in comparison to the MRI features. The APOE-ε4 allele, the best-known genetic risk factor for AD, has always been chosen alongside the other two APOE alleles (ε2 and ε3). The first section of Discussion provides a detailed analysis of the identified genetic features shown in Figure 2.

*DAT score distribution and accuracy across different stratified groups*

Figure 3 displays the DAT score distribution pattern for all stratified groups using genetic, MRI, and genetic+MRI features. GENDATS (Figure 3, top row) shows concentration below the threshold for sNC and uNC and above the threshold for the rest of the groups, while MRDATS and MRGENDATS (Figure 3, middle and bottom row) indicate concentration below the threshold for sNC, uNC, sMCI, and pNC and above the threshold for the rest. The majority of the sMCI group (purple) was misclassified using genetic-only features and the majority of the pNC (orange) groups were misclassified using the MRI-only features. Combining both features resulted in a DAT score distribution that was neutral for sMCI and pNC, while it showed improved performance for the rest.

Figure 4 displays the classification accuracy achieved by comparing a subject's actual diagnosis (DAT-: sNC, uNC, and sMCI, DAT+: pNC, pMCI, eDAT, and sDAT) with the DAT+/DAT- class labels obtained using the 0.5 threshold. For most groups, combining MRI and genetic data yielded better accuracy results than using either feature alone. The exceptions were the sMCI and pNC groups. The sMCI group had a low accuracy (0.356) when genetic features were used, but a higher accuracy when MRI features were used (0.614). The pNC group, on the other hand, had a low accuracy (0.143) using MRI features, but a very high accuracy (0.8571) using genetic features. Combining features for these two groups yielded an accuracy that was in between the two DAT scores trained with individual features.

*Classification accuracy across NC/MCI/DAT groups and DAT-/DAT+ classes*

Figure 5, left column, compares the classification accuracy of the conventional NC, MCI, and DAT groups using our three feature types. Here, NC is made up of the sNC, uNC, and pNC stratified groups, MCI is made up of the sMCI and pMCI groups, and DAT is made up of the eDAT and sDAT groups. Using the combined data (MRI+genetic) resulted in better performance than using either feature alone in all groups. When compared to MRI data, genetic data produced slightly better results for the NC group (0.824 vs. 0.809), whereas MRI data produced better results for the MCI (0.609 vs. 0.516) and DAT (0.866 vs. 0.803) groups. Overall, NC and DAT groups had higher classification accuracy in comparison to MCI.

Figure 5, right column, compares the classification accuracy of the DAT- and DAT+ classes. The DAT- class includes the sNC, uNC, and sMCI stratified groups, whereas the DAT+ class includes the pNC, pMCI, eDAT, and sDAT groups. For the DAT- class, MRI data had the highest accuracy and genetic data had the lowest accuracy, whereas for the DAT+ class, the combined data (MRI+genetic) had the highest accuracy and MRI data had the lowest accuracy. Overall, the accuracy for both DAT- and DAT+ classes appears to be in the same range.

*DAT score distribution among training and testing groups*

Figure 6 shows the histogram distribution of the DAT score among the training groups (sNC and sDAT) for each data type. All three histograms show substantial distinction between the DAT+ (blue) and DAT- (green) classes with the MRGENDATS histogram (Figure 6, bottom row) showing the best performance. The mean DAT score for the sNC group (the smaller the better) decreased from 0.257 and 0.208 with genetic-only or MRI-only features respectively, to 0.119 using the combined features. The mean DAT score for the sDAT group (the larger the better) has increased from 0.717 and 0.777 with genetic-only or MRI-only features respectively, to 0.845 with the combined features.

Figure 7, displays the distribution of the DAT score among the unseen testing groups (uNC, pNC, sMCI, pMCI, and eDAT) for each feature type. The DAT scores for genetic features (GENDATS, Figure 7, top row) demonstrated a higher concentration in the middle while MRDATS and MRGENDATS (Figure 7, middle and bottom rows) show a better separation between the DAT- and DAT+ classes. The mean GENDATS values were between 0.4 and 0.7 for all subgroups with uNC and sMCI (DAT- groups) having slightly smaller values than the rest. The mean MRDATS and MRGENDATS values for DAT- groups (uNC and sMCI) were smaller than DAT+ groups (pMCI and eDAT) with the exception of pNC group.

*Comparison between feature selection methods*

Figure 8 shows the top 17 features selected using LASSO and Fisher/t-test and their corresponding selection frequency. The color map indicates the frequency of selection (in percent) for each of the features using the ensemble classifier. For example, 80% means that a particular feature has been selected using 8 of the F=10 classifiers in the ensemble. Because of the fundamental differences between the feature selection methods, the top 17 features are different for each feature type. For MRI, 12 features have been selected using either Welch's t-test or LASSO (Figure 8, Right column), while 6 genetic features have been mutually selected by Fisher's exact test and LASSO (Figure 8, Left column). The sparsity of the selected genetic features can be explained by the large number of initial features.

Table 4 compares the classification performance of LASSO and Fisher/t-test (using Fisher's exact test on genetic and Welch's t-test on MRI data) methods using genetic, MRI, and genetic+MRI features. In both training and testing phases, the Fisher/t-test method outperformed LASSO when the combination of MRI and genetic data were used, with the only exception that LASSO was slightly higher but non-significant specificity on testing subjects (Fisher/t-test: 0.579 ± 0.021 and LASSO: 0.587 ± 0.03). Using genetic+MRI features, significantly better results (with 0.01 and 0.05 p-values) were obtained with the Fisher/t-test method. Using genetic-only features, Fisher-based feature selection resulted in statistically better training performance, but there was no clear winner in the testing results. T-test-based feature selection gave a slightly better test performance when only MRI features were used. Overall, the Fisher/t-test method outperformed LASSO in most cases.

*GENDATS results using known AD-related SNPs in literature*

Figure 9 displays the DAT score distribution among the 7 stratified groups using the above two SNP sets and our SNP set (extracted using Fisher's exact test) as features. The GENDATS distribution using SNPs from Giri et al. [54] and the Alzgene database (top and middle rows, respectively) is highly concentrated around the 0.5 threshold for all 7 stratified groups. For most groups, the median value is also really close to the threshold, indicating a very small difference between the GENDATS distribution of the stratified groups and a random-like prediction pattern using those SNPs. GENDATS distribution using our method (bottom row), on the other hand, shows a clear distinction between different stratified groups. Using our method, the sNC and uNC groups had a pattern similar to DAT-, while the rest of the groups had a pattern similar to DAT+. When the AD-related SNPs from the literature were used, the sNC and uNC groups showed a similar pattern to DAT-, but no conclusion can be drawn for the rest of the stratified groups, which show no clear tendency to either DAT- or DAT+.

The classification accuracy of the 7 stratified groups using the above two SNP sets and our SNP set is shown in Figure 10. As can be seen, accuracy for most of the stratified groups is around 0.5 when using the SNP sets from Giri et al. [54] and the Alzgene database (blue and cyan, respectively), indicating a random-like pattern in prediction. The sNC and uNC groups appear to have slightly better classification accuracy than the other groups using the SNP sets from the literature, but they still perform worse when compared to our method (yellow). The sMCI group has a low accuracy using our method, indicating that sMCI subjects have a similar pattern to DAT+ rather than DAT-, but no conclusion can be drawn for the sMCI group using SNPs from the literature because the accuracy is very close to 0.5. Overall, using SNPs selected using our method as genetic features yielded better results than AD-related SNPs previously reported in the literature.

## DISCUSSION

*Analyzing genetic discoveries*

Our results replicated some of the AD related genes reported in the literature suggesting the effectiveness of our method. In addition, we identified potentially novel SNPs that could be further explored to verify their associations with DAT. Table 5 includes information about SNPs that have been selected at least twice (Selection frequency ≥ 20% in Figure 2). Features with similar frequency of selection were considered to have the same level of importance, and therefore, Table 5 shows those SNPs that were not included in the top 17 features as well. The rs1864036, rs12522102, and rs17197559 SNPs belong to an uncharacterized RNA gene on chromosome 5 called LOC105379004 and were selected 50%, 30%, and 20% of the time respectively. To date, there is no existing knowledge of the relationship between these SNPs and AD, warranting further investigation. The rs4953672 located between the HAAO and MTA3 genes (chromosome 2), rs2085925 on gene TRAPPC9 (chromosome 8), and rs6116375 on gene PRNP (chromosome 20) and were selected 50%, 40% and 30% of the time respectively. These genes have been reported in previous studies to be associated with AD, brain tissue development or degeneration, and mental disorder [43–47]. Our study has revealed 8 novel SNPs. Four of these SNPs are on chromosome X, three on the LOC105379004 gene on chromosome 5, and one on chromosome 6, indicating the potential importance of chromosomes X and 5 in the development or progression of Alzheimer's Disease.

*Feature selection and combination methods*

In designing the DAT scores, we selected only the pertinent input features by using feature selection methods that were most appropriate for our data type. To avoid overfitting, we restricted the number of features to 10% of the total number of training data (174), yielding 17 features. To evaluate this setup, we ran additional tests with a different number of features (for example, 10 or 25 features instead of 17 for each MRI and genetic feature set) and tried different feature combination methods such as ranking and varying ratios, but they either degraded performance or did not result in a statistically better result. In addition, to evaluate the performance of our data-type-specific feature selection procedure, we compared our method with a regression-based feature selection method called LASSO [40] which showed the superior performance of our method.

*Comprehensive analysis of DAT score distribution for the stratified groups*

In order to interpret the DAT score distribution results (Figure 3) accurately, it is necessary to be mindful about the difference between the characteristics of MRI and the genetic features. Genetic features remain almost the same over time and are not dependent on the longitudinal changes, while MRI features are highly time sensitive and can change drastically over time. A subject may have multiple MRI visits during the study window therefore carrying additional longitudinal information. However, only the baseline MRI imaging data for each subject was included in this study, indicating only the subject's current clinical diagnosis.

The GENDATS value for the sMCI group fell above the threshold for the majority of the subjects suggesting that based on the genetic data, sMCI subjects have a similar pattern to DAT rather than NC. GENDATS for the rest of the stratified groups followed our anticipated pattern. MRDATS for the pNC group was highly concentrated below the threshold and had a similar distribution to the sNC and uNC groups. These similarities can be explained by referring to the fact that only the baseline MRI data has been used and all of these three groups are in a healthy condition at baseline. We anticipate that incorporating longitudinal MRI data can help improve the results for the pNC group. MRDATS followed our expected pattern for the other stratified groups.

MRGENDATS had a higher concentration on the correct side of the threshold for those groups that have previously been classified correctly using MRI and genetic features. MRGENDATS for sMCI was concentrated below the threshold which indicates adding MRI data to genetic may prevent misclassification due to the sole use of genetic data. On the other hand, adding genetic data to MRI for the pNC group has resulted in an almost even MRGENDATS distribution (6 subjects were above and 8 subjects were below the threshold, while only 2 subjects were above the threshold for MRDATS). Possible explanations include lower prediction power of genetic data or small number of subjects in the pNC category. This may be addressed in the future by changing the ratio of the final selected MRI and genetic features and increasing the sample size of the data.

*Benefits of combining MRI and genetic data for DAT prediction*

MRI and genetic data have unique characteristics and can contribute to DAT prediction in different ways. MRI data can reveal anatomical changes in the brain, whereas genetic data can be used to assess the risk of developing DAT even before the symptoms appear. As shown in Figure 4, adding MRI data to genetic data can improve the prediction accuracy for subjects in the sMCI group, and adding genetic data to MRI data can improve the prediction accuracy for subjects in the pNC groups. According to Figure 4, when a single data modality failed to correctly predict the outcome, adding another data modality with distinct properties showed to be beneficial in boosting the performance. When both modalities were successful in predicting the outcome, combining them produced a more accurate result than using either modality alone.

The main focus of our study and key novel contribution is to compare the relative performance of 1) SNP-based genotype features, 2) MRI-based phenotype features, and 3) combined genotype and phenotype features using comprehensive feature selection and aggregation methods. We conducted our experiments by adopting previously validated and published classification and feature selection methods [4]. Because the study's goal is not to develop new classification methods with the highest classification performance, obtaining the highest absolute accuracy of the machine-learning methods was not considered the main focus of the experiments reported here. Rather, since the goal of this experiment was to evaluate the relative performance comparison of different features selected from different modalities (e.g. genotype, phenotype, and genotype+phenotype), the results are reported as found.

To this end, we have performed the above feature comparison using our novel cohort stratification which considers future progression for all individuals in identifying them to individual subgroups yielding some very challenging but also very consequential subgroups (such as pNC – individuals who are currently cognitively healthy but are known to later develop AD). Our main conclusion from the study is: 1) for most cohorts, combining MRI and genetic data yields better accuracy results than using either feature set alone, but, 2) for specific subpopulations such as sMCI and pNC, one modality is found to dominate, for example, genotype features perform better for pNC detection vis-à-vis phenotype features for sMCI. Hence, we showed that a naïve feature concatenation approach is likely insufficient and this finding highlights the importance for further studies to develop smartly-weighted multi-modal feature aggregation using novel information fusion and machine learning methods.

*Application of DAT score in a clinical setting*

In this study, a stratified scheme was used to further break down the standard NC, MCI and DAT categories into smaller groups that take into account the longitudinal diagnosis of a subject and provide a clinically relevant perspective. We trained our network on subjects belonging to the extreme ends of the DAT spectrum (sNC and sDAT) enabling the opportunity to effectively learn distinction between healthy and AD patterns with the highest possible degree of certainty. The trained model was then used to predict a quantitative biomarker based on MRI, genetic, and MRI+genetic features. The only information needed to generate these predictions is that extracted from MRI and genetic data, and the model does not need to have access to the clinical diagnosis. In a clinical setting, clinicians can use our trained model to predict a quantitative score indicating the similarity between a subject's observed pattern based on MRI and genetic data at the time of clinical visit and AD patterns. This will help predict whether the subjects belong to the DAT- (non-progressive) or DAT+ (progressive) categories, which is extremely useful at the MCI stage in identifying those who will progress to AD in the future. We have previously conducted independent validation on real clinical samples using a similar method on FDG-PET data to enable the translation of these methods and test their usefulness in clinical practice [55].

*Analyzing the statistical significance of pNC results*

The sample size of the pNC stratified subgroup is relatively smaller compared to other subgroups. This is a result of our novel stratification method, which classifies subjects based on their past, present, and future longitudinal disease progressions. We have performed rigorous statistical tests to analyze the statistical significance of our evaluation results on pNC subjects (n=14). When comparing GENDATS and MRDATS results, we look at the same group of 14 patients and classify them using two different classifiers, one using genetic data and the other using MRI data. Because we are looking at the same patients, we have used a paired t-test evaluating the difference between the same patient under classifier #1 (based on GENDATS) and classifier #2 (based on

MRDATS). Our test statistic is the following:

$$t = \frac{\bar{Y}_1 - \bar{Y}_2}{\sqrt{\frac{\widehat{\sigma_1^2}}{n} + \frac{\widehat{\sigma_2^2}}{n}}} \quad (1)$$

where $\bar{Y}_1 - \bar{Y}_2$ indicates the mean difference between pairs of measurements in the two classifiers, $\widehat{\sigma_1^2}$ and $\widehat{\sigma_2^2}$ are the variances, and $n$ is the population size. We utilized a one-sided test to check if the predicted DAT score of classifier #1 is significantly greater than the DAT score predicted by classifier #2, and our results showed significantly ($t(13) = 4.33$, $p = 0.0004$) improved predictive power for the GENDATS (0.64 ± 0.07) compared to the MRDATS (0.24 ± 0.05).

It is important to note that the population size (n=14) is taken into account in the test (equation (1)). Thus, despite the relatively small population size, we can detect a significant difference between the two classifiers. The small population size usually leads to big variance estimates which can make the two classifiers hard to distinguish. However, here the variance estimates are small enough, even considering the small population size, to conclude that the two classifiers are significantly different.

*Comparison with previous imaging genetics studies*

In this study, we used all available SNPs in the ADNI database. The main advantage of using all SNPs is that it allows us to investigate potential novel genetic risk factors along with our main task of future DAT prediction. One drawback of using high-dimensional data is that it may contain information that is irrelevant to the task [48]. To avoid this problem and to ensure a strong association between the selected features and the disease pattern, we have implemented an extensive feature selection method in three steps, as previously discussed in Methods, while many previous studies have limited their SNPs to those on the top AD gene candidates according to the Alzgene database (www.alzgene.org) [14, 17, 19–22], and others chose their top SNPs based on the findings of previous studies in the literature [15, 16].

An et al. [14] have proposed a hierarchical feature and sample selection method for AD diagnosis using MRI and SNP data and evaluated their method using conventional binary classification tasks. For DAT vs. NC task (ours sDAT vs. sNC), using only SNP data they received Acc=77.6% and AUC=85.5% (ours: Acc=81.9% and AUC=90.3%) and for MRI+SNP they got Acc=92.4% and AUC=97.4% (ours: Acc=92% and AUC=98%). Our method outperformed theirs when using genetic features, with a 4% increase in both Accuracy and AUC, and had a comparable performance when using combined features. Zhou et al. [21] have proposed a stage-wise deep learning algorithm for AD prediction using MRI, PET, and SNP data and evaluated their method using traditional classification tasks. Their multiclass classification showed a median accuracy of less than 55% while we achieved 63.2% accuracy using MRI and SNP data. Our training (sNC/sDAT) results were slightly better than their NC/DAT results (Acc: 92% vs 91.7%) despite the fact that we have not used PET data. Venugopala et al. [18] have utilized deep learning methods to investigate the effects of combining multimodal data such as MRI, genetic, and clinical data on AD prediction. They performed a binary classification task on NC vs DAT/MCI (ours sNC vs. sDAT), using only MRI data, they received Acc=86% (ours: Acc=88.2%), using SNP information only, they got an accuracy of 89% (ours: Acc= 81%) and for MRI+SNP, their best performing model received Acc=75% (ours: Acc=92%). Our method outperformed theirs when using MRI by more than 2%, and MRI+SNP by 17%, while their network performed better when using only SNP data (8% percent higher accuracy). Zhang et al. [20] have studied the effects of combining MRI, SNP, CSF and PET modalities on AD prediction by performing conventional classification tasks using linear support vector machines (SVMs) and three intrinsic feature selection algorithms. Their best performing model had a classification accuracy of 94.8% for AD vs. NC task using 378 features from all 4 modalities which is slightly better than our sNC vs. sDAT accuracy of 92% using only 34 SNP and MRI features.

*Limitations and future direction*

Our study has some limitations. Our results are limited by the small sample size selected from the ADNI dataset and their characteristics, especially for the pNC and eDAT stratified groups. We have conducted our analysis based on the information available in the ADNI study window. Subjects currently on the DAT-trajectory might receive a follow- up diagnosis of DAT in the future. In that case, we will review our network's prediction in the future and investigate if the misclassification for those subjects can be justified by the fact that they did not have a DAT follow-up diagnosis. Another approach to addressing this limitation is to limit the follow-up duration to a specific time, such as 3 or 5 years after the baseline, and study the probability of pathological onset during that time-frame, which is a clinically relevant approach and is similar to survival analysis for Alzheimer's disease.

In this study, we have achieved our primary goal of predicting the future conversion to Alzheimer's disease by extracting MRI and genetic information from sNC and sDAT stratified groups as they represent the DAT- (stable) and DAT+ (progressive) categories with the highest degree of certainty. We have previously used this technique to develop a) fluorodeoxyglucose positron emission tomography (FDG-PET) imaging-based score [24] and b) MRI-based score [4] for early DAT detection and achieved the state-of-the-art performance. However, this approach could reflect a predisposed bias for processes that are not necessarily linked to AD. A potential solution to this problem could be incorporating subjects at earlier stage of the AD progression (e.g. sMCI, and pMCI group) to learn potential patterns that are caused by additional factors during early stage of the AD pathogenesis.

As a part of our future work, we plan to: 1) use the UK Biobank database (https://www.ukbiobank.ac.uk), a large-scale biomedical database including genotype data from 500,000 participants and brain MRI data from over 44,000 participants, to increase the sample size, especially for subjects in the pNC and eDAT stratified groups, and to construct a more robust model, 2) evaluate the generalizability of the genetic features discovered in this study by training our model with subjects from a different AD-related dataset, 3) expand our methodology and use deep-learning- based models such as fully connected networks and Deep Embedding network to investigate the potential improvement of the classification performance, 4) use longitudinal MRI data instead of the baseline to include the time related changes, 5) limit the follow-up duration to a certain time and incorporate survival analysis approaches and compare the results with the current design, 6) remove individual heterogeneity due to age in addition to other factors by means of GLM for MRI data, 7) incorporate smart nonlinear approaches that can capture the importance of each MRI and genetic feature to better integrate them.

## ACKNOWLEDGEMENTS

Funding for this research is gratefully acknowledged from Alzheimer Society Research Program, National Science Engineering Research Council (NSERC), Canadian Institutes of Health Research (CIHR), Fondation Brain Canada, Pacific Alzheimer's Research Foundation, the Michael Smith Foundation for Health Research (MSFHR), the National Institute on Aging (R01 AG055121-01A1, R01 AG069765-01, R01 AG071514-01), National Institute of Neurological Disorders and Stroke (NINDS) (R01 NS101483- 01A1), and Precision Imaging Beacon, University of Nottingham. We thank Compute Canada for the computational infrastructure provided for the data processing in this study.

Data collection and sharing for this project was funded by the Alzheimer's Disease Neuroimaging Initiative (ADNI) (National Institutes of Health Grant U01 AG024904) and DOD ADNI (Department of Defense award number W81XWH-12-2-0012). ADNI is funded by the National Institute on Aging, the National Institute of Biomedical Imaging and Bioengineering, and through generous contributions from the following: AbbVie, Alzheimer's Association; Alzheimer's Drug Discovery Foundation; Araclon Biotech; BioClinica, Inc.; Biogen; Bristol-Myers Squibb Company; CereSpir, Inc.; Cogstate; Eisai Inc.; Elan Pharmaceuticals, Inc.; Eli Lilly and Company; EuroImmun; F. Hoffmann-La Roche Ltd and its affiliated company Genentech, Inc.; Fujirebio; GE Healthcare; IXICO Ltd.; Janssen Alzheimer Immunotherapy Research & Development, LLC.; Johnson & Johnson Pharmaceutical Research & Development LLC.; Lumosity; Lundbeck; Merck & Co., Inc.; Meso Scale Diagnostics, LLC.; NeuroRx Research; Neurotrack Technologies; Novartis Pharmaceuticals Corporation; Pfizer Inc.; Piramal Imaging; Servier; Takeda Pharmaceutical Company; and Transition Therapeutics. The Canadian Institutes of Health Research is providing funds to support ADNI clinical sites in Canada. Private sector contributions are facilitated by the Foundation for the National Institutes of Health (www.fnih.org). The grantee organization is the Northern California Institute for Research and Education, and the study is coordinated by the Alzheimer's Therapeutic Research Institute at the University of Southern California. ADNI data are disseminated by the Laboratory for Neuro Imaging at the University of Southern California.

## CONFLICT OF INTEREST

The authors have no conflict of interest to report.

## References

[1] Alzheimer's Association (2018) 2018 Alzheimer's disease facts and figures. *Alzheimer's Dement* **14**, 367–429.

[2] He Y, Wang L, Zang Y, Tian L, Zhang X, Li K, Jiang T (2007) Regional coherence changes in the early stages of Alzheimer's disease: A combined structural and resting-state functional MRI study. *Neuroimage* **35**, 488–500.

[3] Hua X, Leow AD, Parikshak N, Lee S, Chiang MC, Toga AW, Jack CR, Weiner MW, Thompson PM (2008) Tensor-based morphometry as a neuroimaging biomarker for Alzheimer's disease: An MRI study of 676 AD, MCI, and normal subjects. *Neuroimage* **43**, 458–469.

[4] Popuri K, Ma D, Wang L, Beg MF (2020) Using machine learning to quantify structural MRI neurodegeneration patterns of Alzheimer's disease into dementia score: Independent validation on 8,834 images from ADNI, AIBL, OASIS, and MIRIAD databases. *Hum Brain Mapp* **41**, 4127–4147.

[5] Scheltens P, Leys D, Barkhof F, Huglo D, Weinstein HC, Vermersch P, Kuiper M, Steinling M, Wolters EC, Valk J (1992) Atrophy of medial temporal lobes on MRI in" probable" Alzheimer's disease and normal ageing: diagnostic value and neuropsychological correlates. *J Neurol Neurosurg Psychiatry* **55**, 967–972.

[6] Vemuri P, Jack CR (2010) Role of structural MRI in Alzheimer's disease. *Alzheimers Res Ther* **2**, 1–10.

[7] Lambert J-C, Ibrahim-Verbaas CA, Harold D, Naj AC, Sims R, Bellenguez C, Jun G, DeStefano AL, Bis JC, Beecham GW, others (2013) Meta-analysis of 74,046 individuals identifies 11 new susceptibility loci for Alzheimer's disease. *Nat Genet* **45**, 1452–1458.

[8] Bertram L, McQueen MB, Mullin K, Blacker D, Tanzi RE (2007) Systematic meta-analyses of Alzheimer disease genetic association studies: The AlzGene database. *Nat Genet* **39**, 17–23.

[9] Jansen IE, Savage JE, Watanabe K, Bryois J, Williams DM, Steinberg S, others (2019) Genome-wide meta-analysis identifies new loci and functional pathways influencing Alzheimer's disease risk. *Nat Genet* **51**, 404–413.

[10] Karch CM, Goate AM (2015) Alzheimer's disease risk genes and mechanisms of disease pathogenesis. *Biol Psychiatry* **77**, 43–51.

[11] Kunkle BW, Grenier-Boley B, Sims R, Bis JC, Damotte V, Naj AC, Boland A, Vronskaya M, Van Der Lee SJ, Amlie-Wolf A, others (2019) Genetic meta-analysis of diagnosed Alzheimer's disease identifies new risk loci and implicates Aβ, tau, immunity and lipid processing. *Nat Genet* **51**, 414–430.

[12] Schwartzentruber J, Cooper S, Liu JZ, Barrio-Hernandez I, Bello E, Kumasaka N, Young AMH, Franklin RJM, Johnson T, Estrada K, others (2021) Genome-wide meta-analysis, fine-mapping and integrative prioritization implicate new Alzheimer's disease risk genes. *Nat Genet* 1–11.

[13] Sumirtanurdin R, Thalib AY, Cantona K, Abdulah R (2019) Effect of genetic polymorphisms on alzheimer's disease treatment outcomes: An update. *Clin Interv Aging* **14**, 631–642.

[14] An L, Adeli E, Liu M, Zhang J, Lee SW, Shen D (2017) A Hierarchical Feature and Sample Selection Framework and Its Application for Alzheimer's Disease Diagnosis. *Sci Rep* **7**, 1–11.

[15] Biffi A, Anderson CD, Desikan RS, Sabuncu M, Cortellini L, Schmansky N, Salat D, Rosand J (2010) Genetic variation and neuroimaging measures in Alzheimer disease. *Arch Neurol* **67**, 677–685.

[16] Ning K, Chen B, Sun F, Hobel Z, Zhao L, Matloff W, Toga AW (2018) Classifying Alzheimer's disease with brain imaging and genetic data using a neural network framework. *Neurobiol Aging* **68**, 151–158.

[17] Peng J, An L, Zhu X, Jin Y, Shen D (2016) Structured sparse kernel learning for imaging genetics based Alzheimer's disease diagnosis. In *International Conference on Medical Image Computing and Computer-Assisted Intervention*, pp. 70–78.

[18] Venugopalan J, Tong L, Hassanzadeh HR, Wang MD (2021) Multimodal deep learning models for early detection of Alzheimer's disease stage. *Sci Rep* **11**, 1–13.

[19] Wang H, Nie F, Huang H, Risacher SL, Saykin AJ, Shen L (2012) Identifying disease sensitive and quantitative trait-relevant biomarkers from multidimensional heterogeneous imaging genetics data via sparse multimodal multitask learning. *Bioinformatics* **28**, i127–i136.

[20] Zhang Z, Huang H, Shen D (2014) Integrative analysis of multi-dimensional imaging genomics data for alzheimer's disease prediction. *Front Aging Neurosci* **6**, 260.

[21] Zhou T, Thung KH, Zhu X, Shen D (2019) Effective feature learning and fusion of multimodality data using stage-wise deep neural network for dementia diagnosis. *Hum Brain Mapp* **40**, 1001–1016.

[22] Zhou T, Liu M, Thung KH, Shen D (2019) Latent Representation Learning for Alzheimer's Disease Diagnosis with Incomplete Multi-Modality Neuroimaging and Genetic Data. *IEEE Trans Med Imaging* **38**, 2411–2422.

[23] Mueller SG, Weiner MW, Thal LJ, Petersen RC, Jack CR, Jagust W, Trojanowski JQ, Toga AW, Beckett L (2005) Ways toward an early diagnosis in Alzheimer's disease: the Alzheimer's Disease Neuroimaging Initiative (ADNI). *Alzheimer's Dement* **1**, 55–66.

[24] Popuri K, Balachandar R, Alpert K, Lu D, Bhalla M, Mackenzie IR, Hsiung RGY, Wang L, Beg MF (2018) Development and validation of a novel dementia of Alzheimer's type (DAT) score based on metabolism FDG-PET imaging. *NeuroImage Clin* **18**, 802–813.

[25] Chang CC, Chow CC, Tellier LCAM, Vattikuti S, Purcell SM, Lee JJ (2015) Second-generation PLINK: Rising to the challenge of larger and richer datasets. *Gigascience* **4**.

[26] Saykin AJ, Shen L, Foroud TM, Potkin SG, Swaminathan S, Kim S, Risacher SL, Nho K, Huentelman MJ, Craig DW, others (2010) Alzheimer's Disease Neuroimaging Initiative biomarkers as

quantitative phenotypes: Genetics core aims, progress, and plans. *Alzheimer's Dement* **6**, 265–273.

[27] Dale AM, Fischl B, Sereno MI (1999) Cortical surface-based analysis: I. Segmentation and surface reconstruction. *Neuroimage* **9**, 179–194.

[28] Desikan RS, Ségonne F, Fischl B, Quinn BT, Dickerson BC, Blacker D, Buckner RL, Dale AM, Maguire RP, Hyman BT, Albert MS, Killiany RJ (2006) An automated labeling system for subdividing the human cerebral cortex on MRI scans into gyral based regions of interest. *Neuroimage* **31**, 968–980.

[29] Fischl B, Salat DH, Busa E, Albert M, Dieterich M, Haselgrove C, Van Der Kouwe A, Killiany R, Kennedy D, Klaveness S, others (2002) Whole brain segmentation: automated labeling of neuroanatomical structures in the human brain. *Neuron* **33**, 341–355.

[30] Ma D, Popuri K, Bhalla M, Sangha O, Lu D, Cao J, Jacova C, Wang L, Beg MF (2019) Quantitative assessment of field strength, total intracranial volume, sex, and age effects on the goodness of harmonization for volumetric analysis on the ADNI database. *Hum Brain Mapp* **40**, 1507–1527.

[31] Dietterich TG (2000) Ensemble methods in machine learning. In *International workshop on multiple classifier systems*, pp. 1–15.

[32] Akritas MG, Politis DN (2003) *Recent Advances and Trends in Nonparametric Statistics*, Elsevier.

[33] Loew MH (2000) Feature extraction. *Handb Med imaging* **2**, 273–342.

[34] Sonka M, Fitzpatrick JM (2000) Handbook of medical imaging (Volume 2, Medical image processing and analysis). In *SPIE- The international society for optical engineering*.

[35] Raamana PR, Weiner MW, Wang L, Beg MF (2015) Thickness network features for prognostic applications in dementia. *Neurobiol Aging* **36**, S91–S102.

[36] Brownlee J How to Choose a Feature Selection Method For Machine Learning. *Mach Learn Mastery.* https://machinelearningmastery.com/feature-selection-with-real-and-categorical-data/, Last updated August 20, 2020, Accessed on August 23, 2021.

[37] Fisher RA, others (1934) Statistical methods for research workers. *Stat methods Res Work*.

[38] Welch BL (1947) The generalization of student's' problem when several different population variances are involved. *Biometrika* **34**, 28–35.

[39] Sullivan GM, Feinn R (2012) Using Effect Size—or Why the P Value Is Not Enough. *J Grad Med Educ* **4**, 279–282.

[40] Tibshirani R (1996) Regression Shrinkage and Selection Via the Lasso. *J R Stat Soc Ser B* **58**, 267–288.

[41] Damoulas T, Girolami MA (2008) Probabilistic multi-class multi-kernel learning: On protein fold recognition and remote homology detection. *Bioinformatics* **24**, 1264–1270.

[42] Cristianini N, Shawe-Taylor J (2004) *Kernel methods for pattern analysis*, Cambridge University Press Cambridge.

[43] Devyatkin VA, Redina OE, Kolosova NG, Muraleva NA (2020) Single-Nucleotide Polymorphisms Associated with the Senescence-Accelerated Phenotype of OXYS Rats: A Focus on Alzheimer's Disease-Like and Age-Related-Macular-Degeneration-Like Pathologies. *J Alzheimers Dis* **73**, 1167–1183.

[44] Golanska E, Hulas-Bigoszewska K, Sieruta M, Zawlik I, Witusik M, Gresner SM, Sobow T, Styczynska M, Peplonska B, Barcikowska M, Liberski PP, Corder EH (2009) Earlier onset of alzheimer's disease: Risk polymorphisms within PRNP, PRND, CYP46, and APOE genes. *J Alzheimer's Dis* **17**, 359–368.

[45] Mochida GH, Mahajnah M, Hill AD, Basel-Vanagaite L, Gleason D, Hill RS, Bodell A, Crosier M, Straussberg R, Walsh CA (2009) A Truncating Mutation of TRAPPC9 Is Associated with Autosomal-Recessive Intellectual Disability and Postnatal Microcephaly. *Am J Hum Genet* **85**, 897–902.

[46] Potts RC, Zhang P, Wurster AL, Precht P, Mughal MR, Wood WH, Zhang Y, Becker KG, Mattson MP, Pazin MJ (2011) CHD5, a brain-specific paralog of Mi2 chromatin remodeling enzymes, regulates expression of neuronal genes. *PLoS One* **6**, e24515.

[47] Sutphin GL (2018) SYSTEMIC ELEVATION OF 3-HYDROXYANTHRANILIC ACID (3HAA) TO EXTEND LIFESPAN AND DELAY ALZHEIMER'S PATHOLOGY. *Innov Aging* **2**, 74–74.

[48] Yu L, Liu H (2003) Efficiently handling feature redundancy in high-dimensional data. In *Proceedings of the ACM SIGKDD International Conference on Knowledge Discovery and Data Mining*, pp. 685–690.

[49] Meda SA, Narayanan B, Liu J, Perrone-Bizzozero NI, Stevens MC, Calhoun VD, Glahn DC, Shen L, Risacher SL, Saykin AJ, Pearlson GD (2012) A large scale multivariate parallel ICA method reveals novel imaging-genetic relationships for Alzheimer's disease in the ADNI cohort. *Neuroimage* **60**, 1608–1621.

[50] Vounou M, Janousova E, Wolz R, Stein JL, Thompson PM, Rueckert D, Montana G (2012) Sparse reduced-rank regression detects genetic associations with voxel-wise longitudinal phenotypes in Alzheimer's disease. *Neuroimage* **60**, 700–716.

[51] Humphries CE, Kohli MA, Nathanson L, Whitehead P, Beecham G, Martin E, Mash DC, Pericak-Vance MA, Gilbert J (2015) Integrated whole transcriptome and DNA methylation analysis identifies gene networks specific to late-onset Alzheimer's disease. *J Alzheimer's Dis* **44**, 977–987.

[52] Gal J, Chen J, Katsumata Y, Fardo DW, Wang WX, Artiushin S, Price D, Anderson S, Patel E, Zhu H, Nelson PT (2018) Detergent insoluble proteins and inclusion body-like structures immunoreactive for PRKDC/DNA-PK/DNA-PKcs, FTL, NNT, and AIFM1 in the amygdala of cognitively impaired elderly persons. *J Neuropathol Exp Neurol* **77**, 21–39.

[53] Young FB, Butland SL, Sanders SS, Sutton LM, Hayden MR (2012) Putting proteins in their place: palmitoylation in Huntington disease and other neuropsychiatric diseases. *Prog Neurobiol* **97**, 220–

238.

[54] Giri M, Zhang M, Lu¨ Y (2016) Genes associated with Alzheimer's disease: an overview and current status. *Clin Interv Aging* **11**, 665-681.

[55] Ma D, Yee E, Stocks JK, Jenkins LM, Popuri K, Chausse G, Wang L, Probst S, Beg MF (2021) Blinded Clinical Evaluation for Dementia of Alzheimer's Type Classification Using FDG-PET: A Comparison Between Feature-Engineered and Non-Feature-Engineered Machine Learning Methods. *J Alzheimer's Dis* **80**, 715–726.

| Dementia trajectory | Group name | Clinical diagnosis at imaging | Clinical progression | Subjects [M:F] | Age[c] [Years] | CSF[a,c] [t-tau/A$\beta_{1-42}$] |
|---|---|---|---|---|---|---|
| DAT-[b] | sNC: stable NC | NC[a] | **NC** → **NC** | 58:51 | 75.79 (4.93) | 0.34 (0.23) |
| DAT- | uNC: unstable NC | NC | **NC** → MCI | 14:8 | 76.57 (3.70) | 0.39 (0.19) |
| DAT- | sMCI: stable MCI | MCI[a] | NC → **MCI** *or* **MCI** → **MCI** | 65:36 | 74.70 (7.35) | 0.67 (0.52) |
| DAT+[b] | pNC: progressive NC | NC | **NC** → MCI → DAT | 6:8 | 76.49 (4.33) | 0.75 (0.42) |
| DAT+ | pMCI: progressive MC | MCI | NC → **MCI** → DAT *or* **MCI** → DAT | 99:56 | 73.85 (6.85) | 0.82 (0.45) |
| DAT+ | eDAT*: early DAT | DAT[a] | NC → MCI → **DAT** *or* MCI → **DAT** | 2:2 | 75.80 (4.13) | 0.65 (0.00) |
| DAT+ | sDAT: stable DAT | DAT | **DAT** → **DAT** | 74:64 | 75.19 (7.54) | 0.89 (0.46) |

[a] NC: normal controls, MCI: mild cognitive impairment, DAT: dementia of Alzheimer's type
CSF: cerebrospinal fluid, t-tau: total tau, A$\beta_{1-42}$: beta amyloid 1-42
[b] DAT+: On DAT trajectory, i.e., at some point in time, these subjects will be clinically diagnosed as DAT
DAT-: not on the DAT trajectory and will not get a DAT diagnosis in the ADNI window
[c] The mean (standard deviation) age and CSF measure values within each group are given
CSF measures were only available for a subset of images in each of the groups:
sNC (57), uNC (17), sMCI (55), pNC (8), pMCI (88), eDAT (1), sDAT (87)
* The eDAT group has a small sample size (4 subjects), but it has been included in the study for the sake of completeness.

Table 1: Stratification of ADNI subjects based on their longitudinal clinical diagnosis. The stratification was based on two criteria, clinical diagnosis of subjects at the time of MRI image acquisition and their longitudinal clinical progression. Each subject is assigned a membership in the form of 'prefixGroup', where 'Group' is the clinical diagnosis at the current imaging visit, and 'prefix' signals past or future clinical diagnoses. For e.g., a subject is designated as pNC if the subject was assigned an NC diagnosis at that particular imaging visit, but the subject converts to DAT at a future timepoint. The eDAT images are associated with the diagnosis of DAT, but the subject had received NC or MCI status during previous ADNI visits (conversion within ADNI window), whereas, the sDAT images belong to the subjects with a consistent clinical diagnosis of DAT throughout the ADNI study window, hence these individuals have progressed to DAT prior to their ADNI recruitment. Clinical diagnosis at the time of imaging is shown in **bold** under the "Clinical progression" column.

| Giri *et al.* [54]<br>SNP ID (gene) | Alzgene<br>SNP ID (gene) |
|---|---|
| rs10498633 (SLC24A4/RIN3) | $\varepsilon 4$ (APOE) |
| rs17125944 (FERMT2) | $\varepsilon 3$ (APOE) |
| rs3851179 (PICALM) | rs3851179 (PICALM) |
| rs541458 (PICALM) | rs541458 (PICALM) |
| rs610932 (MS4A6A) | rs610932 (MS4A6A) |
| rs3865444 (CD33) | rs3865444 (CD33) |
| rs3826656 (CD33) | rs3826656 (CD33) |
| rs670139 (MS4A4E) | rs670139 (MS4A4E) |
| rs9296559 (CD2AP) | rs9296559 (CD2AP) |
| rs3764650 (ABCA7) | rs3764650 (ABCA7) |
| rs7561528 (BIN1) | rs7561528 (BIN1) |
| rs744373 (BIN1) | rs744373 (BIN1) |
| rs2718058 (NME8) | rs12989701 (BIN1) |
| rs3818361 (CR1) | rs3818361 (CR1) |
| rs2305421 (ADAM10) | rs6701713 (CR1) |
| rs11771145 (EPHA1) | rs1408077 (CR1) |
| rs11767557 (EPHA1) | rs11136000 (CLU) |

Table 2: Known AD-related SNPs reported in the literature. Left column: 17 AD-related SNPs reported in Giri et al. [54], Right column: 17 SNPs from the top 10 AD-related genes reported in the Alzgene database. 11 SNPs have been mutually reported to be related with Alzheimer's Disease in both datasets.

| ROI name | Description | Selection frequency (%) [left\|right] |
|---|---|---|
| *Hippocampus* | Allocortex (Subcortical) region | 100 \| 100 |
| *Amygdala* | Subcortical region | 100 \| 100 |
| *Entorhinal* | Cortical region | 100 \| 100 |
| Fusiform | Cortical region | 100 \| 100 |
| Inferior temporal | Cortical region | 100 \| 100 |
| Middle temporal | Cortical region | 100 \| 100 |
| Para hippocampal | Cortical region | 100 \| 70 |
| Inferior parietal | Cortical region | 60 \| 100 |
| Inferior–lateral-ventricle | Inferior or temporal horn of the lateral ventricle | 70 \| 30 |
| Supramarginal | Cortical region | 40 \| 00 |
| Precuneus | Cortical region | 10 \| 20 |

Table 3: Most discriminative MRI features determined by the feature selection process and their frequency of selection. These MRI features indicate the volumetric measures of the brain ROIs. ROIs are listed in the descending order of their total (left and right) selection frequency.

| | | Fisher/t-test (mean ± sd) | LASSO (mean ± sd) |
|---|---|---|---|
| **Training Results (on sNC and sDAT)** | | | |
| Genetic | AUC | 0.882 ± 0.036 | 0.822 ± 0.062† |
| | Sensitivity | 0.771 ± 0.052 | 0.718 ± 0.057† |
| | Specificity | 0.832 ± 0.094 | 0.768 ± 0.119* |
| | BalAccuracy | 0.801 ± 0.045 | 0.743 ± 0.067† |
| | Accuracy | 0.789 ± 0.038 | 0.733 ± 0.054† |
| MRI | AUC | 0.940 ± 0.025 | 0.954 ± 0.026 |
| | Sensitivity | 0.871 ± 0.026 | 0.880 ± 0.032 |
| | Specificity | 0.918 ± 0.047 | 0.905 ± 0.040 |
| | BalAccuracy | 0.894 ± 0.026 | 0.892 ± 0.030 |
| | Accuracy | 0.885 ± 0.023 | 0.888 ± 0.030 |
| Genetic+MRI | AUC | 0.978 ± 0.011 | 0.969 ± 0.026 |
| | Sensitivity | 0.918 ± 0.020 | 0.910 ± 0.038 |
| | Specificity | 0.932 ± 0.054 | 0.932 ± 0.058 |
| | BalAccuracy | 0.925 ± 0.025 | 0.921 ± 0.033 |
| | Accuracy | 0.922 ± 0.017 | 0.916 ± 0.030 |
| **Testing Results (on uNC, sMCI, pNC, pMCI and eDAT)** | | | |
| Genetic | AUC | 0.555 ± 0.009 | 0.541 ± 0.011† |
| | Sensitivity | 0.640 ± 0.024 | 0.639 ± 0.029 |
| | Specificity | 0.411 ± 0.039 | 0.439 ± 0.033 |
| | BalAccuracy | 0.526 ± 0.013 | 0.539 ± 0.011 |
| | Accuracy | 0.545 ± 0.011 | 0.556 ± 0.011 |
| MRI | AUC | 0.654 ± 0.014 | 0.657 ± 0.011 |
| | Sensitivity | 0.639 ± 0.022 | 0.627 ± 0.013 |
| | Specificity | 0.575 ± 0.029 | 0.560 ± 0.018 |
| | BalAccuracy | 0.607 ± 0.008 | 0.593 ± 0.011† |
| | Accuracy | 0.602 ± 0.011 | 0.588 ± 0.012† |
| Genetic+MRI | AUC | 0.660 ± 0.006 | 0.651 ± 0.010† |
| | Sensitivity | 0.662 ± 0.019 | 0.631 ± 0.014† |
| | Specificity | 0.579 ± 0.021 | 0.587 ± 0.030 |
| | BalAccuracy | 0.620 ± 0.007 | 0.609 ± 0.011* |
| | Accuracy | 0.627 ± 0.008 | 0.613 ± 0.008† |

Table 4: Classification performance comparison among Fisher/t-test and LASSO feature selection methods on genetic, MRI and genetic+MRI feature types. Fisher/t-test: Using Fisher's exact test on genetic features and Welch's t-test on MRI features. Training results show the classification performance of groups with the most certain diagnosis (sNC and sDAT) and the testing results show the classification performance of unseen subjects in the remaining stratified groups (uNC, sMCI, pNC, pMCI and eDAT). The best performance has been highlighted in red. Symbol * denotes the t-test with $p < 0.05$ and † denotes the t-test with $p < 0.01$ as compared to the Fisher/t-test results.

| SNP ID | Selection freq (%) | Chromosome | nearest Gene | status |
|---|---|---|---|---|
| $\varepsilon 4$ | 100 | 19 | APOE | known |
| $\varepsilon 3$ | 100 | 19 | APOE | known |
| $\varepsilon 2$ | 100 | 19 | APOE | known |
| rs4953672 | 50 | 2 | HAAO and MTA3 | known [46, 47] |
| rs1864036 | 50 | 5 | LOC105379004 | novel |
| rs2085925 | 40 | 8 | TRAPPC9 | known [43, 45] |
| rs12522102 | 30 | 5 | LOC105379004 | novel |
| rs6116375 | 30 | 20 | PRNP | known [44] |
| rs2405940 | 30 | X | SHROOM2 | known [49] |
| rs10465385 | 30 | X | LINC02154 | novel |
| rs10924809 | 20 | 1 | CNST | known [43] |
| rs2883782 | 20 | 2 | MYO3B | known [50] |
| rs746947 | 20 | 3 | FRMD4B | known [51] |
| rs10510985 | 20 | 3 | FRMD4B | known [51] |
| rs6773506 | 20 | 3 | FRMD4B | known [51] |
| rs7627954 | 20 | 3 | TNIK | known [52] |
| rs17197559 | 20 | 5 | LOC105379004 | novel |
| rs524410 | 20 | 6 | LOC112267968 | novel |
| rs5918417 | 20 | X | SYTL5 | novel |
| rs5918419 | 20 | X | SYTL5 | novel |
| rs1010616 | 20 | X | ZDHHC15 | known [53] |
| rs12860832 | 20 | X | PASD1 | novel |

Table 5: Detail of SNPs selected at least twice using Fisher exact test in the ensemble. For those SNPs that do not fall exactly on a particular gene, nearest genes have been reported. Status column indicates whether SNPs have been previously reported to be associated with Alzheimer's disease or brain tissue degeneration.

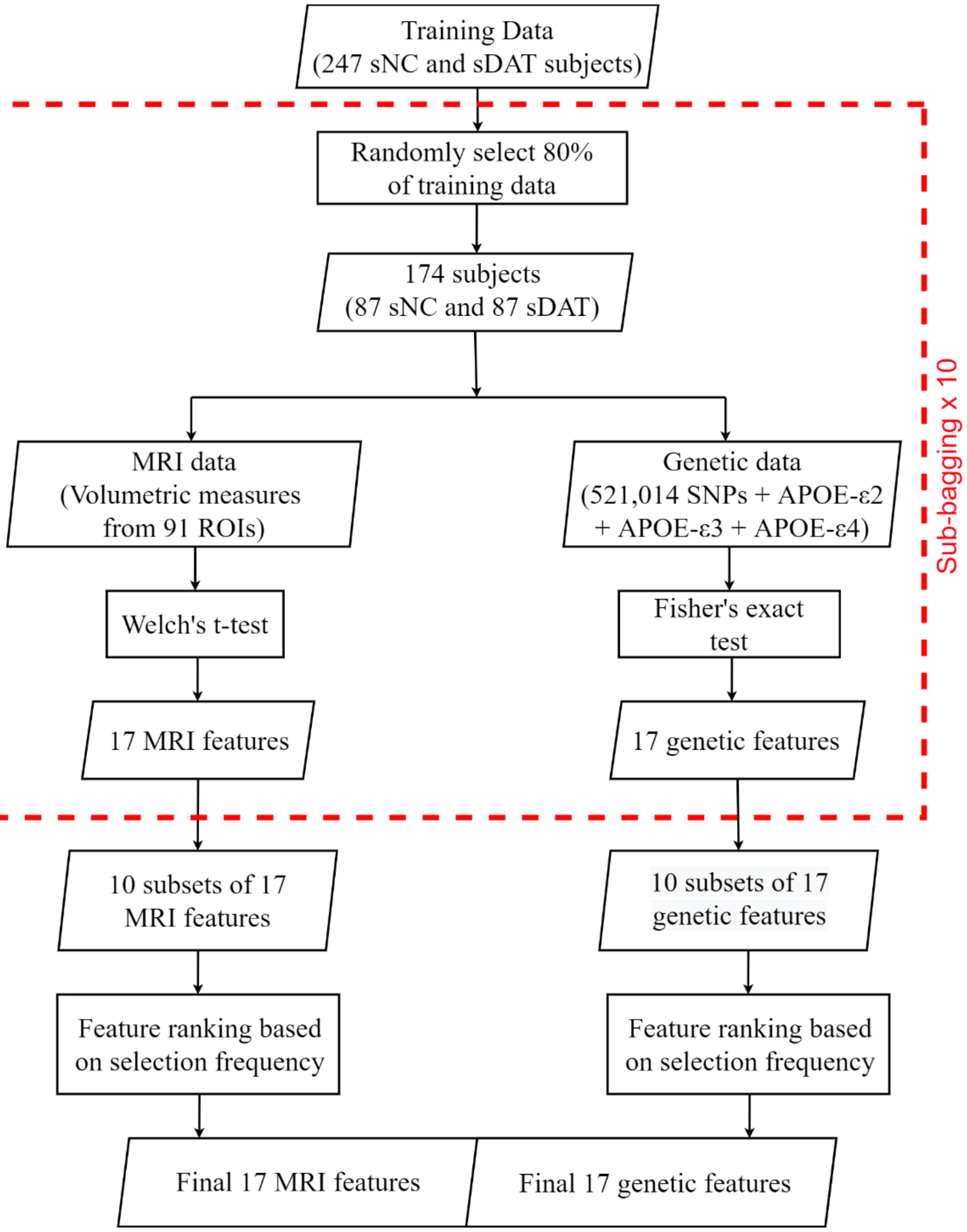

Figure 1: Graphic representation of the feature selection process for MRI and genetic data. Using a sub-bagging approach, F=10 subsets of training data including 80% of the sNC and sDAT subjects are first generated. Then, separate statistical tests are applied on MRI (Welch's t-test) and genetic data (Fisher's exact test) to select the most discriminative k=17 features for each F subsets of data. This process generates 10 sets of k=17 features for each data modality. The features are then ranked based on their selection frequency and the final sets of 17 features are chosen for the DAT score computation step. To investigate the joint effect of MRI and genetic data, features from both modalities were combined (17 + 17 = 34 MRI+genetic features).

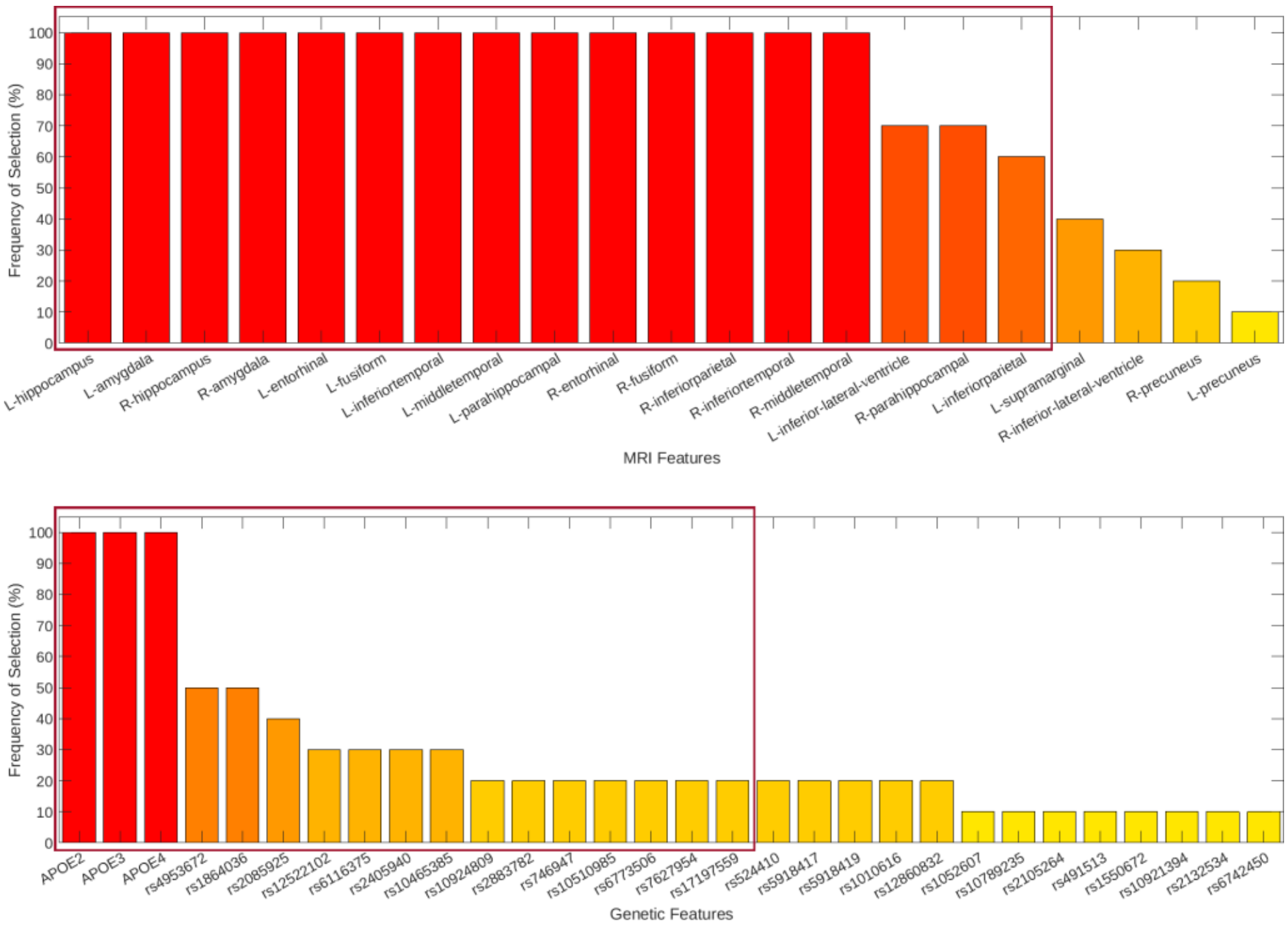

Figure 2: Feature selection results for MRI (top) and genetic (bottom) data. Frequency of selection indicates the amount of time each feature has been selected by the classifiers (For example: 80% means that a particular feature has been selected using 8 of the F=10 classifiers). The top 17 most discriminative features are highlighted for each feature type. Top row: The overall set of MRI features selected by the classifier ensemble using Welch's t-test. L indicates the left hemisphere and R indicates the right hemisphere of the brain. Bottom row: The overall set of genetic features selected by the classifier ensemble using Fisher's exact t-test. Final features (boxed in red) have been chosen randomly if their frequency of selection were the same.

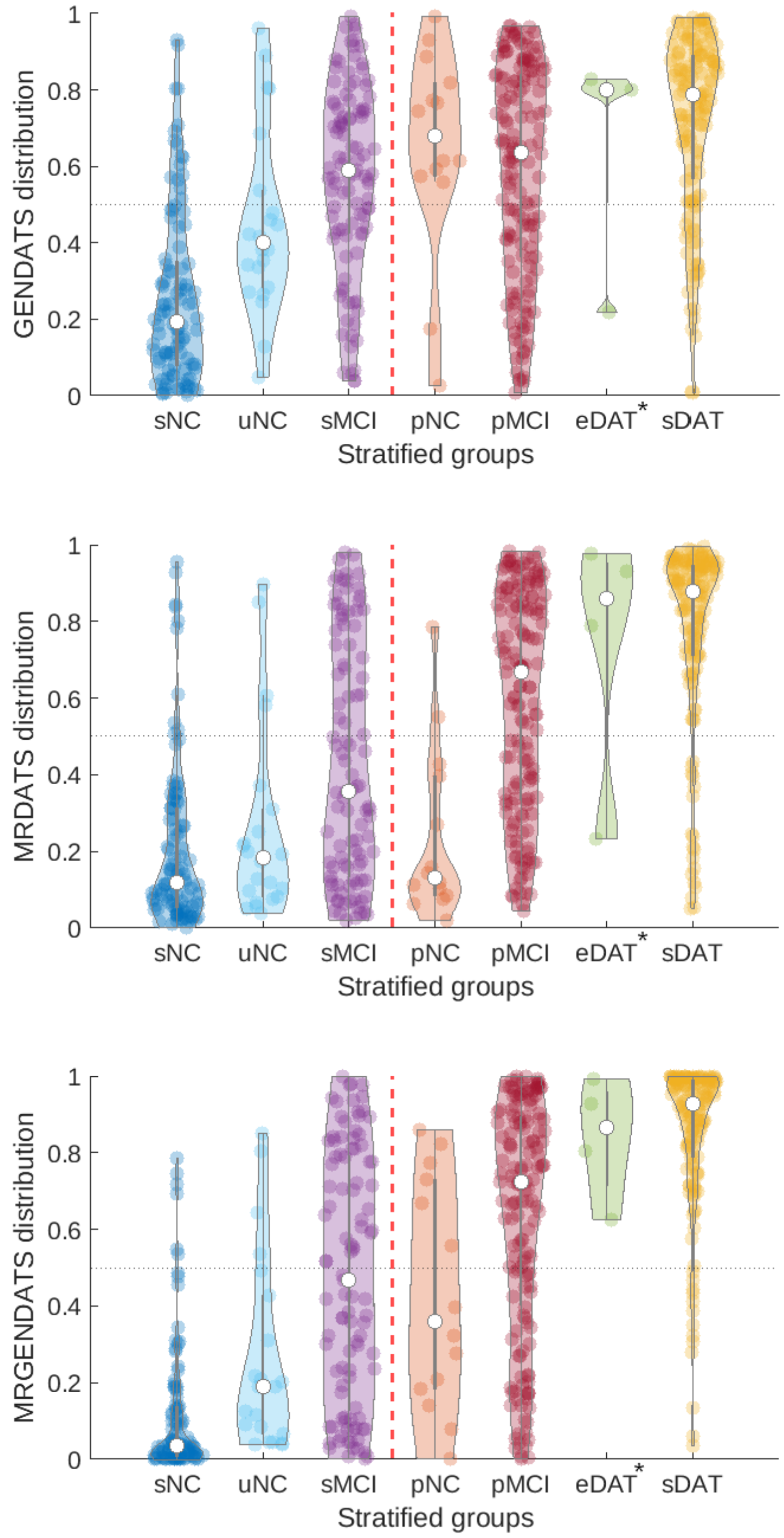

* small sample size (4 subjects)

Figure 3: DAT score distribution among the 7 stratified sub groups for each feature type. sNC, uNC and sMCI groups belong to the DAT- trajectory and pNC, pMCI, eDAT, and sDAT groups belong to the DAT+ trajectory. DAT+/DAT- groups are separated with a red vertical line. A midway threshold of 0.5 for DAT scores is shown using a black horizontal line. Top row: Genetic DAT score (GENDATS) distribution, Middle row: MRI DAT score (MRDATS) distribution, and Bottom row: MRI+Genetic DAT score (MRGENDATS) distribution across different stratified groups. Median of the DAT score for each group is shown using a white circle.

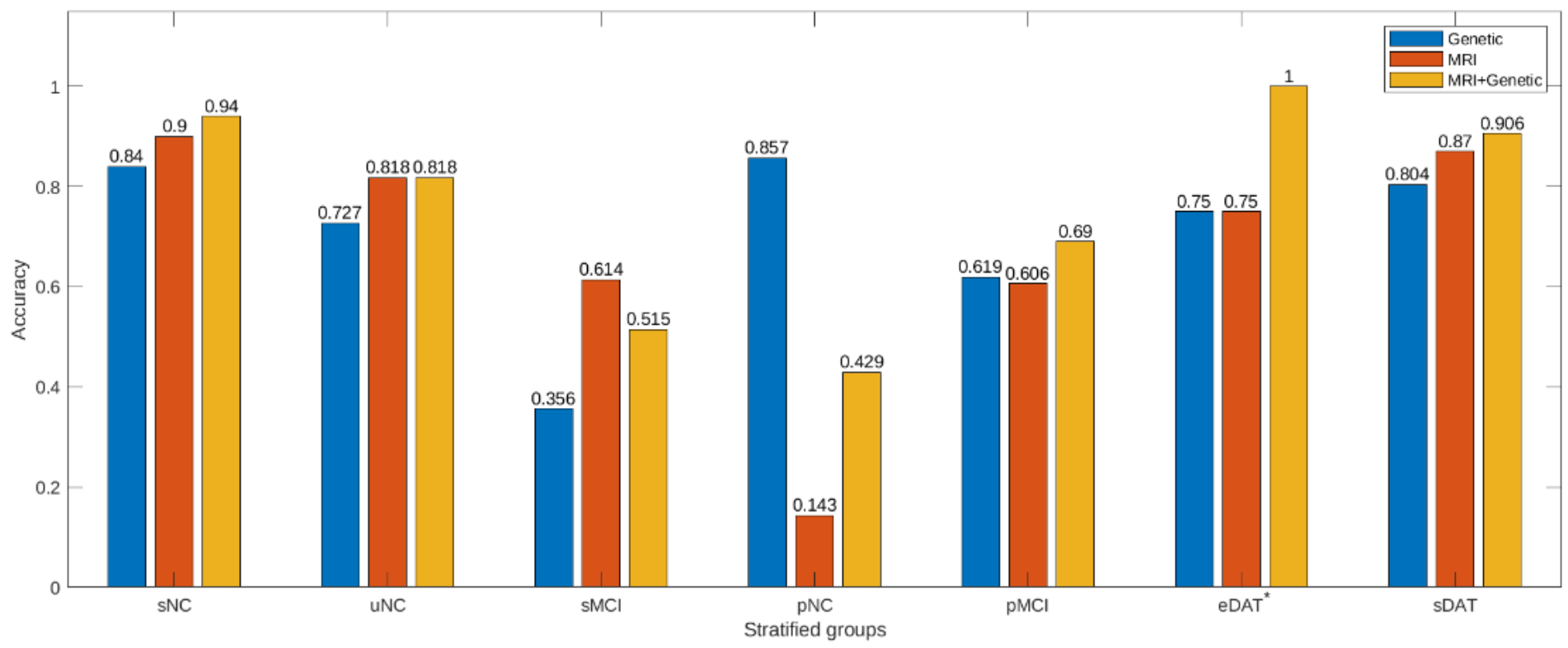

* small sample size (4 subjects)

Figure 4: Classification accuracy for each group obtained by comparing the true diagnostics (DAT-: sNC, uNC, and sMCI, DAT+: pNC, pMCI, eDAT and sDAT) with the dementia trajectories (DAT- or DAT+) assigned to each subject using a threshold from the DAT scores. Blue bars show accuracy using genetic features, orange bars indicate accuracy using MRI features and yellow bars display accuracy for the combined features.

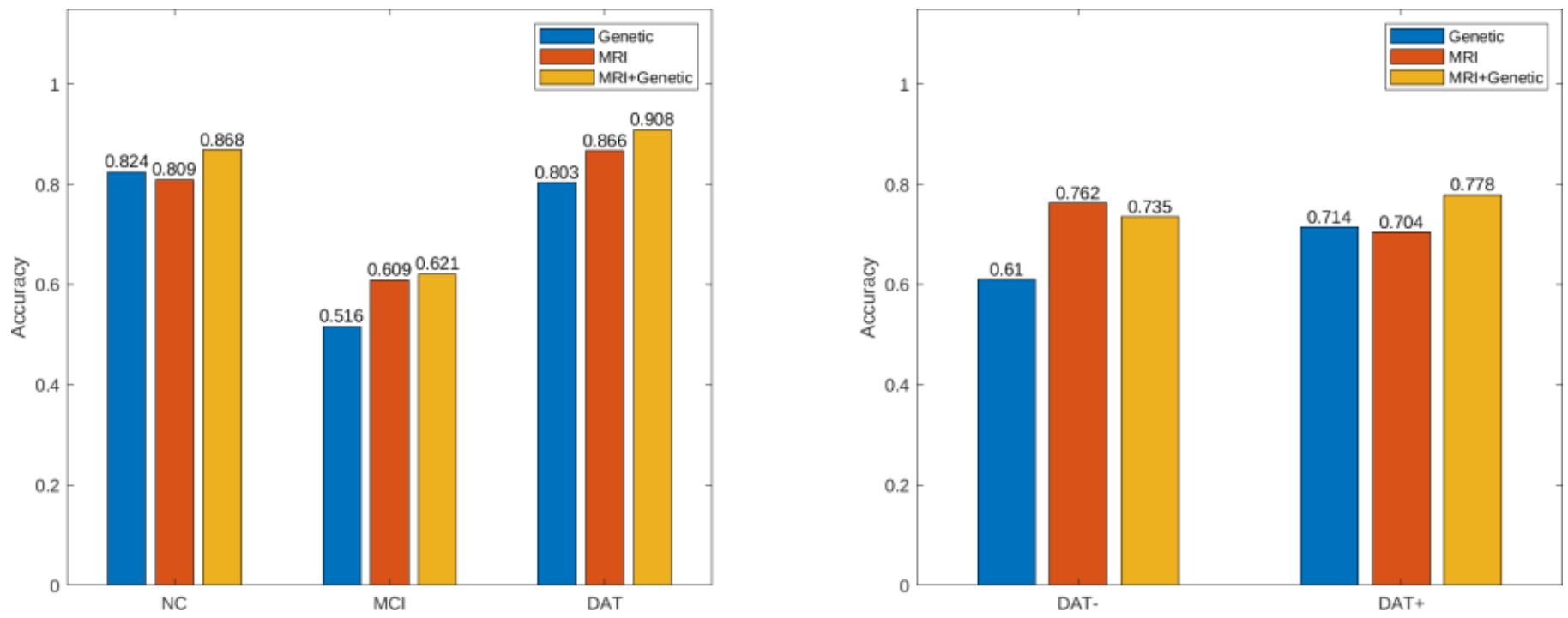

Figure 5: Classification accuracy obtained by comparing the true diagnostics with the dementia trajectories assigned to each subject using a 0.5 threshold from the DAT scores. Left column: classification accuracy comparison between the conventional NC (sNC, uNC, and pNC), MCI (sMCI and pMCI), and DAT (eDAT and sDAT) groups, Right column: classification accuracy comparison between DAT- (sNC, uNC, and sMCI) and DAT+ (pNC, pMCI, eDAT and sDAT) groups. Blue bars show accuracy using genetic features, orange bars indicate accuracy using MRI features and yellow bars display accuracy for the combined features.

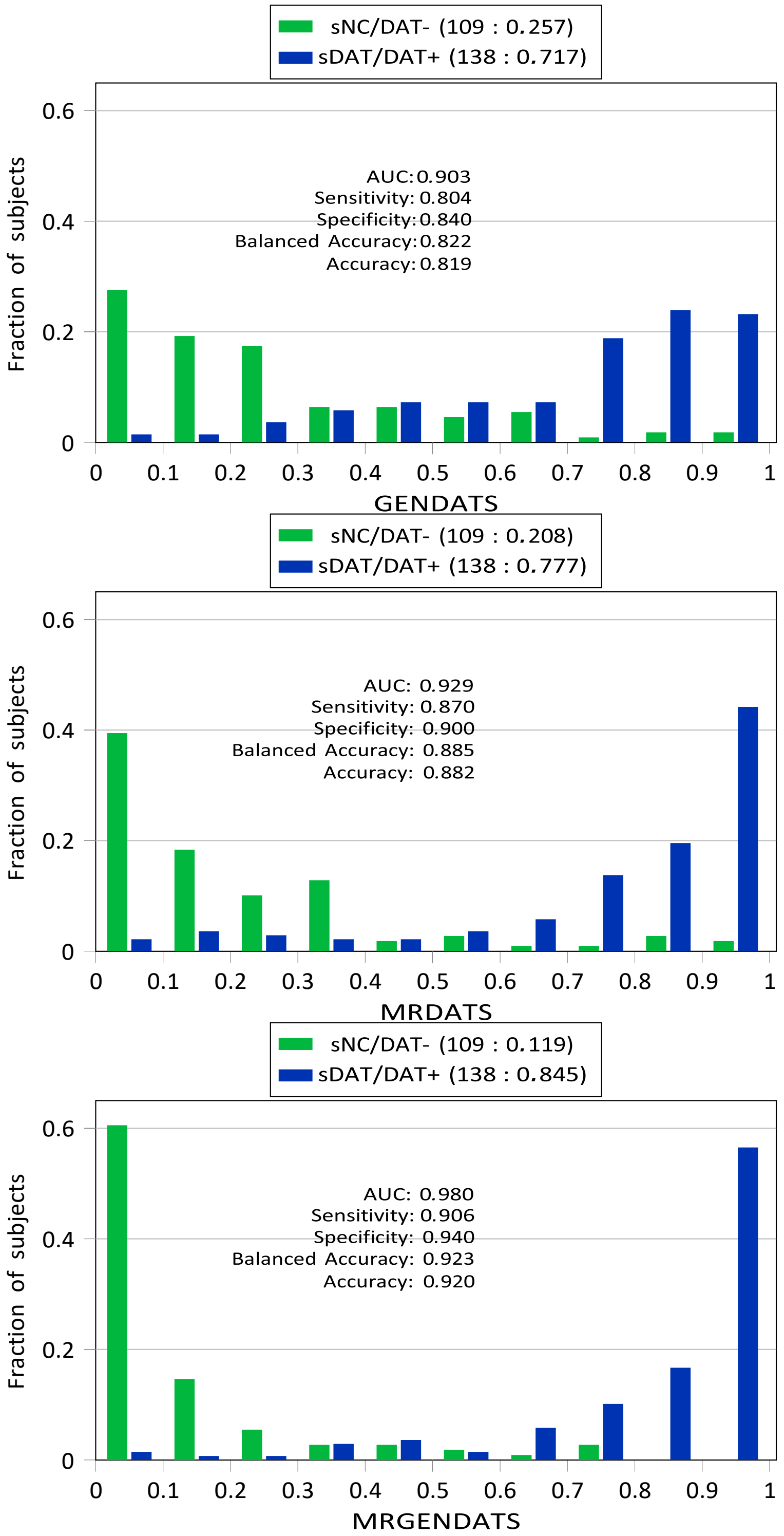

Figure 6: DAT score distribution among sNC and sDAT subjects and classification performance obtained in assigning either the DAT- or DAT+ trajectory using a 0.5 threshold. Top row: Genetic DAT score (GENDATS) results using only genetic features. Middle row: MRI DAT score (MRDATS) results using only MRI features. Bottom row: MRI+Genetic DAT score (MRGENDATS) results using combined features. The (number of subjects: mean DAT score) is shown for each subgroup. Balanced accuracy is the mean of the sensitivity and specificity measures.

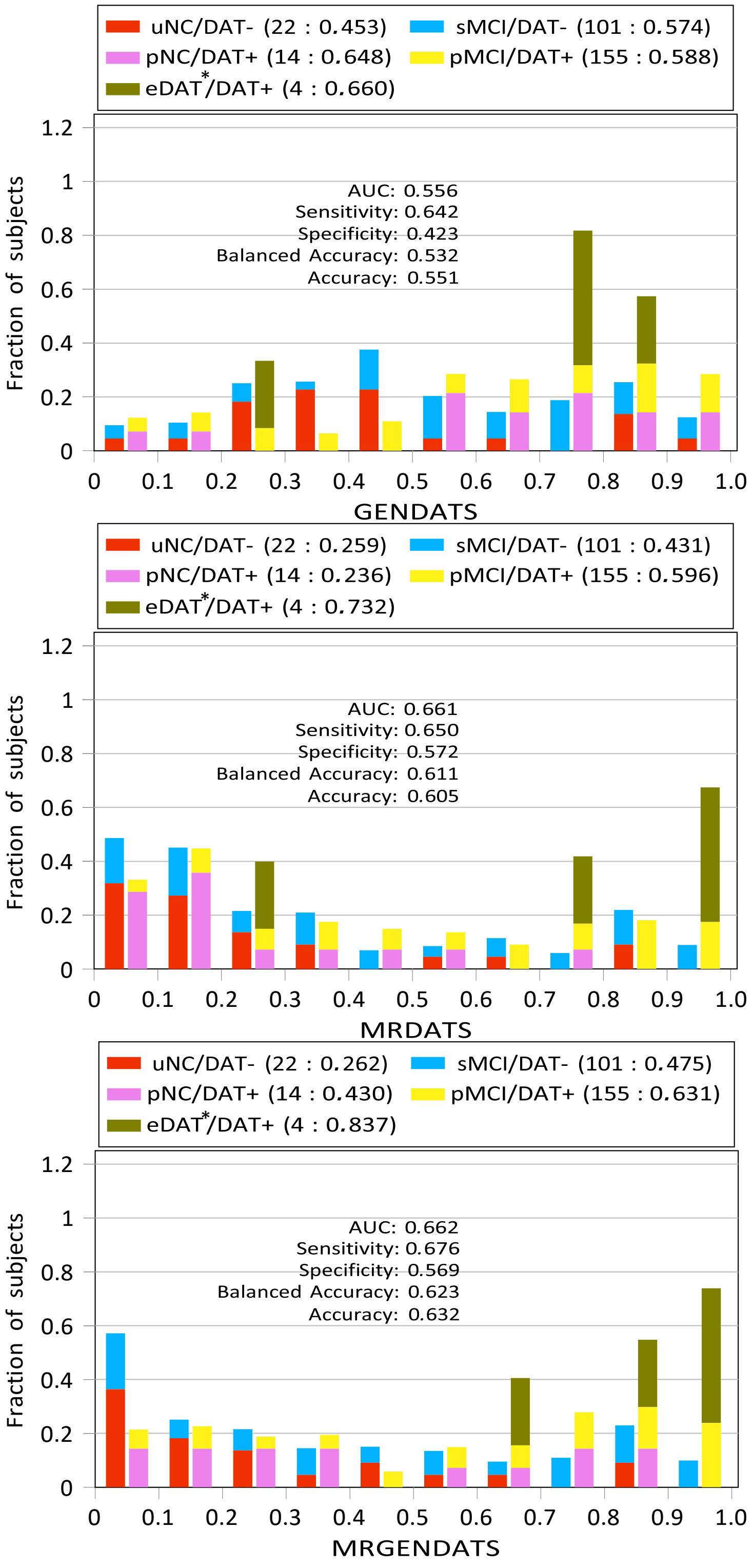

* small sample size (4 subjects)

Figure 7: DAT score distribution among the independent validation subjects. The classification performance was obtained by determining dementia trajectories (DAT- or DAT+) for each subject using a 0.5 threshold. The MRDATS histograms corresponding to the DAT- (uNC, sMCI) and the DAT+ (pNC, pMCI, eDAT) trajectories are stacked together respectively. Top row: Genetic DAT score (GENDATS) results using only genetic features. Middle row: MRI DAT score (MRDATS) results using only MRI features. Bottom row: MRI+Genetic DAT score (MRGENDATS) results using combined features. The (number of subjects: mean DAT score) is shown for each subgroup.

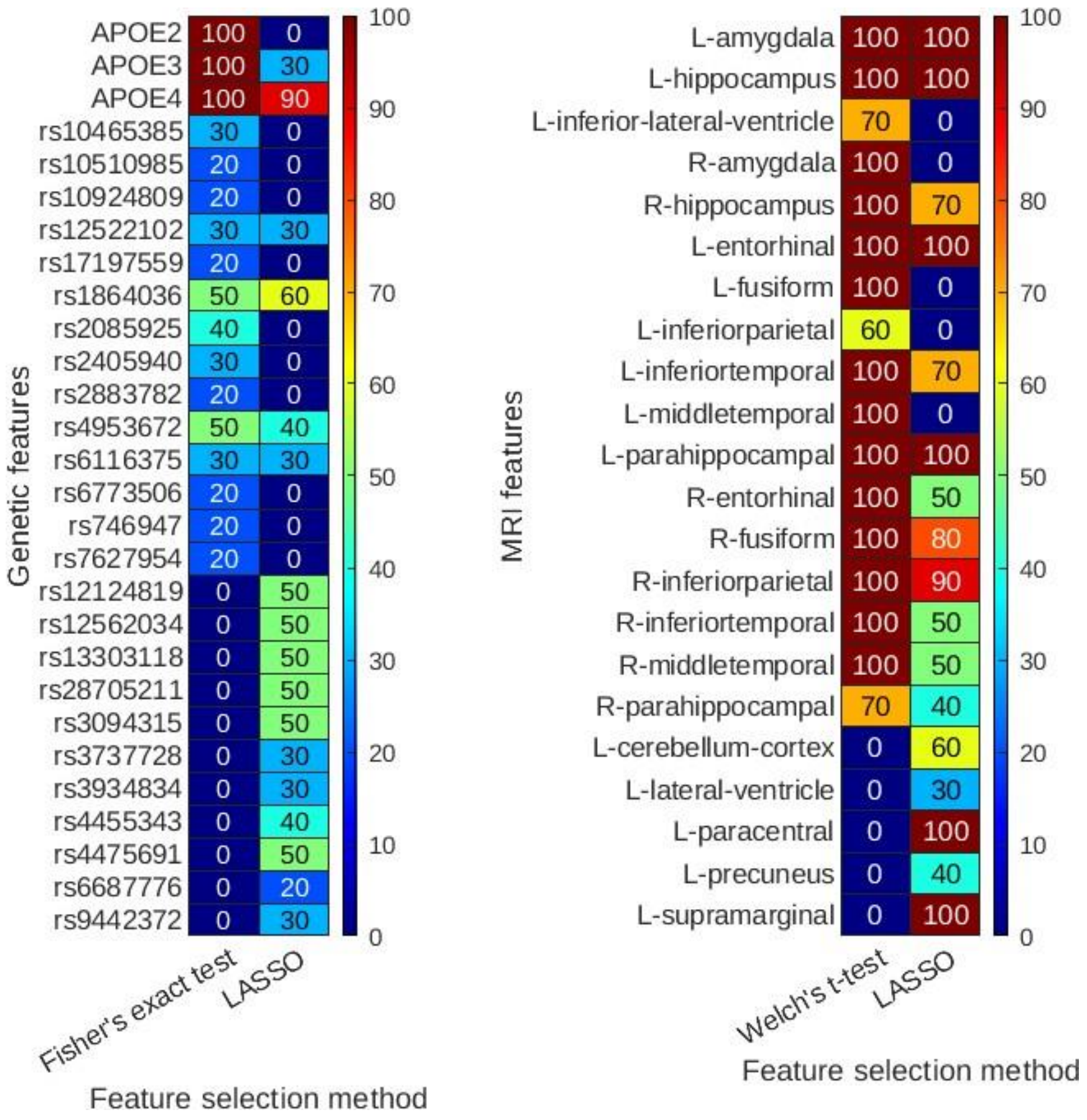

Figure 8: The top 17 MRI and genetic features selected using different feature selection methods are shown. Left column: genetic features selected using Fisher's exact test and LASSO. Right column: MRI features selected using Welch's t-test and LASSO. L indicates the left hemisphere and R indicates the right hemisphere of the brain. The number within each cell indicates the selection frequency (%) for each feature.

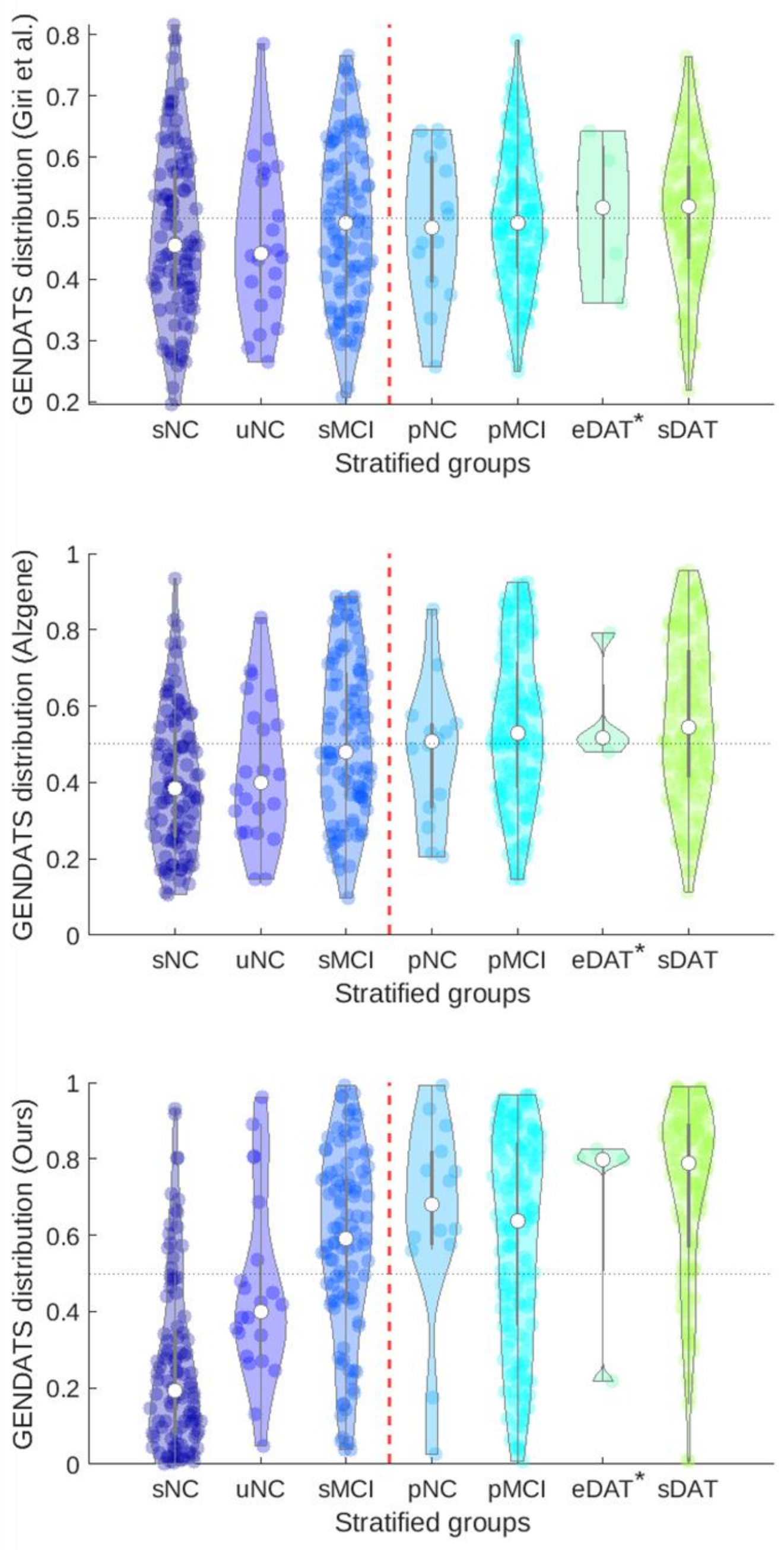

* small sample size (4 subjects)

Figure 9: GENDATS distribution among the 7 stratified sub groups for each SNP set. sNC, uNC and sMCI groups belong to the DAT- trajectory and pNC, pMCI, eDAT, and sDAT groups belong to the DAT+ trajectory. DAT+/DAT- groups are separated with a red vertical line. A midway threshold of 0.5 for DAT scores is shown using a black horizontal line. Top row: GENDATS distribution using SNPs reported in Giri et al. [54], Middle row: GENDATS distribution using SNPs from the top 10 AD-related genes reported in the Alzgene database, and Bottom row: GENDATS distribution using SNPs extracted from all available SNPs in the ADNI database using Fisher's exact test (our method). Median of the DAT score for each group is shown using a white circle.

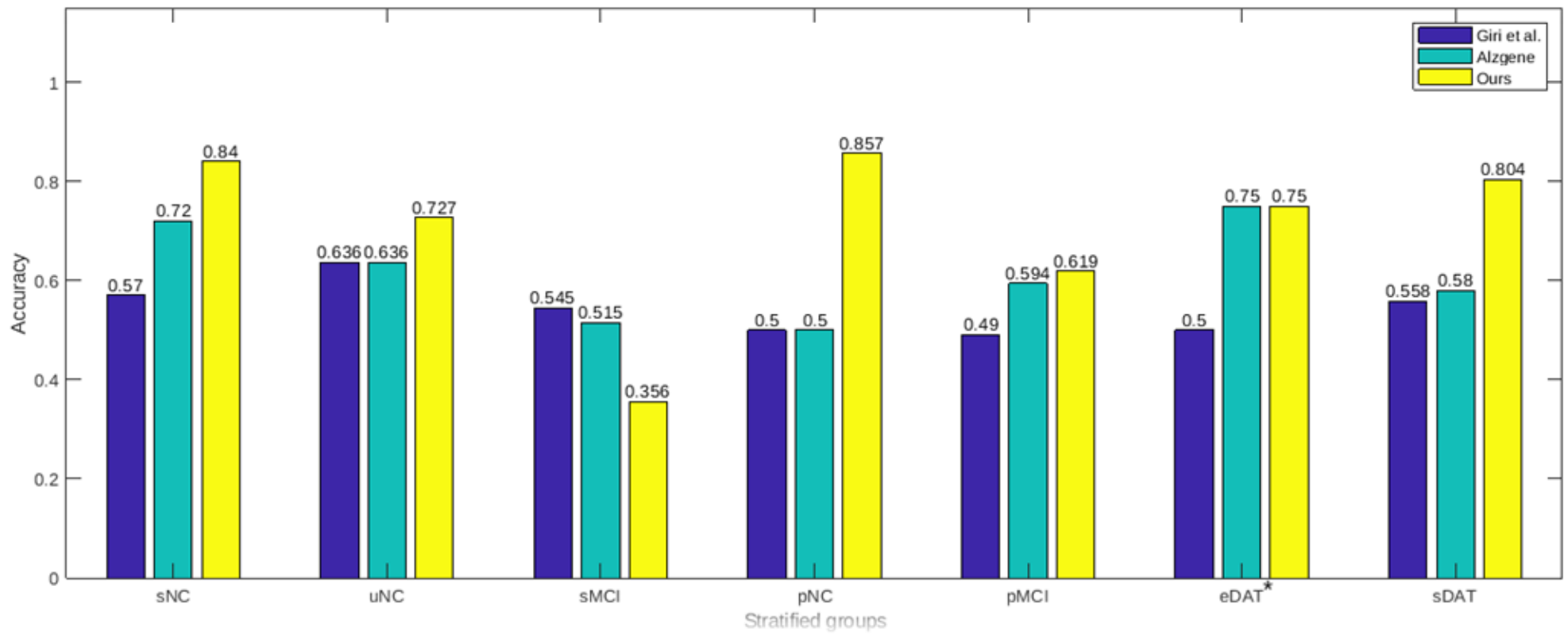

* small sample size (4 subjects)

Figure 10: Classification accuracy for each stratified group obtained by comparing the true diagnostics (DAT-: sNC, uNC, and sMCI, DAT+: pNC, pMCI, eDAT and sDAT) with the dementia trajectories (DAT- or DAT+) assigned to each subject using a 0.5 threshold from the DAT scores. Blue bars show accuracy using SNPs reported in Giri et al. [54], cyan bars indicate accuracy using SNPs reported in the Alzgene database and yellow bars display accuracy using SNPs extracted from ADNI database with Fisher's exact test (our method).